%% file: root.tex
\documentclass[letterpaper, 10 pt, conference, final]{ieeeconf}

\IEEEoverridecommandlockouts                              

\usepackage{amsmath, scalerel} 
\usepackage{amssymb}
\usepackage{amsopn}
\usepackage{graphicx}
\usepackage[style=base]{caption}
\usepackage{subfigure}
\usepackage{color}
\usepackage{quotmark}
\usepackage{booktabs}
\usepackage{todonotes}
\usepackage{lipsum}
\usepackage{import}
\usepackage{footnote}
\usepackage{multirow}
\usepackage{hhline}
\usepackage[pdfborder={0 0 0}, bookmarks=false, final]{hyperref}

\usepackage[abbreviations,binary-units]{siunitx}
\DeclareSIUnit{\mAh}{mAh}
\DeclareSIUnit{\Wh}{Wh}

\usepackage{tikz}
\usetikzlibrary{calc}
\usetikzlibrary{matrix}
\usetikzlibrary{math}
\usetikzlibrary{plotmarks}
\usetikzlibrary{positioning}
\usetikzlibrary{shapes}
\usetikzlibrary{arrows}
\usetikzlibrary{arrows.meta}

\usepackage{pgfplots}
\pgfplotsset{compat=newest}
\usepgfplotslibrary{patchplots}
\usepackage{grffile}

\usepackage{pgfplots}
\pgfplotsset{compat=newest} 
\pgfplotsset{plot coordinates/math parser=false} 
\newlength\figureheight 
\newlength\figurewidth 

\title{\LARGE \bf
Towards Inverse Sensor Mapping in Agriculture
}

\author{Timo Korthals$^{1}$, Mikkel Kragh$^{2}$, Peter Christiansen$^{2}$, and Ulrich R{\"u}ckert$^{1}$
\thanks{$^{1}$Bielefeld University, Cluster of Excellence Cognitive Interaction Technologies, Cognitronics \& Sensor Systems,
        Inspiration 1, 33619 Bielefeld, Germany,
        {\tt\small http://www.ks.cit-ec.uni-bielefeld.de/},
        {\tt\small \{tkorthals, rueckert\}}
        {\tt\small @cit-ec.uni-bielefeld.de}}
\thanks{$^{2}$Aarhus University, Department of Engineering, Finlandsgade 22, DK-8200 Aarhus N, Denmark
        {\tt\small http://eng.au.dk/},
        {\tt\small \{mkha, pech\}}
        {\tt\small @eng.au.dk}}
}

\DeclareMathOperator{\Odd}{Odds}

\DeclareMathOperator*{\Bigcdot}{\scalerel*{\cdot}{\bigodot}}

\begin{document}

\maketitle
\thispagestyle{empty}
\pagestyle{empty}

\begin{abstract}
In recent years, the drive of the Industry 4.0 initiative has enriched industrial and scientific approaches to build self-driving cars or smart factories.
Agricultural applications benefit from both advances, as they are in reality mobile driving factories which process the environment.
Therefore, acurate perception of the surrounding is a crucial task as it involves the goods to be processed, in contrast to standard indoor production lines.
Environmental processing requires accurate and robust quantification in order to correctly adjust processing parameters and detect hazardous risks during the processing.
While today approaches still implement functional elements based on a single particular set of sensors, it may become apparent that a unified representation of the environment compiled from all available information sources would be more versatile, sufficient, and cost effective.
The key to this approach is the means of developing a common information language from the data provided.
In this paper, we introduce and discuss techniques to build so called inverse sensor models that create a common information language among different, but typically agricultural, information providers.
These can be current live sensor data, farm management systems, or long term information generated from previous processing, drones, or satellites.
In the context of Industry 4.0, this enables the interoperability of different agricultural systems and allows information transparency.
\end{abstract}
\section{Introduction}
\label{sec:Introduction}

Agricultural vehicles are complex, mobile processors of biological products that operate in unstructured and constantly changing environment.
While the operation of these vehicles was initially relatively simple, today their setup and use requires trained specialists due to the requirement of increasing efficiency and lowering overall costs.
However, without automation and the augmenting of parameter optimization in the process chain, throughputs, and farming yields would be much smaller than usual.
For instance, automated steering systems employed in harvesting use LiDAR systems to scan the area between the crop and stubble in order to automatically guide the harvester along the edge; and seed drills save GPS data and the machine parameters of sowing which are used later to minimize the utilization of fertilizer spreaders.

Focusing the automation and in particular its implementation, all applications follow the same paradigm of having a distinctive set of sensors, a processing unit, and an actuator interface to steer the vehicle or manipulate process parameters.
While this approach allows simple, distributed and modular modification, with increases in automated functionality its installation and maintenance becomes unfeasible due to the sheer number of sensors and processing units required.
Furthermore, the potential for sensor fusion is completely squandered. 
An alternative approach is pursued by the authors, that of building a common inner semantical representation of the environment based on occupancy grid maps, from which all further automation is derived \cite{Korthals2016, Korthals2016a}.
These grid maps are arranged in multiple overlapping layers, where each one is occupied by localized classifications.

\begin{figure*}
\centering
\begin{scriptsize}
\def\svgwidth{0.75\textwidth}
\begingroup
  \makeatletter
  \providecommand\color[2][]{
    \errmessage{(Inkscape) Color is used for the text in Inkscape, but the package 'color.sty' is not loaded}
    \renewcommand\color[2][]{}
  }
  \providecommand\transparent[1]{
    \errmessage{(Inkscape) Transparency is used (non-zero) for the text in Inkscape, but the package 'transparent.sty' is not loaded}
    \renewcommand\transparent[1]{}
  }
  \providecommand\rotatebox[2]{#2}
  \ifx\svgwidth\undefined
    \setlength{\unitlength}{461.98543519bp}
    \ifx\svgscale\undefined
      \relax
    \else
      \setlength{\unitlength}{\unitlength * \real{\svgscale}}
    \fi
  \else
    \setlength{\unitlength}{\svgwidth}
  \fi
  \global\let\svgwidth\undefined
  \global\let\svgscale\undefined
  \makeatother
  \begin{picture}(1,0.34391988)
    \put(0,0){\includegraphics[width=\unitlength,page=1]{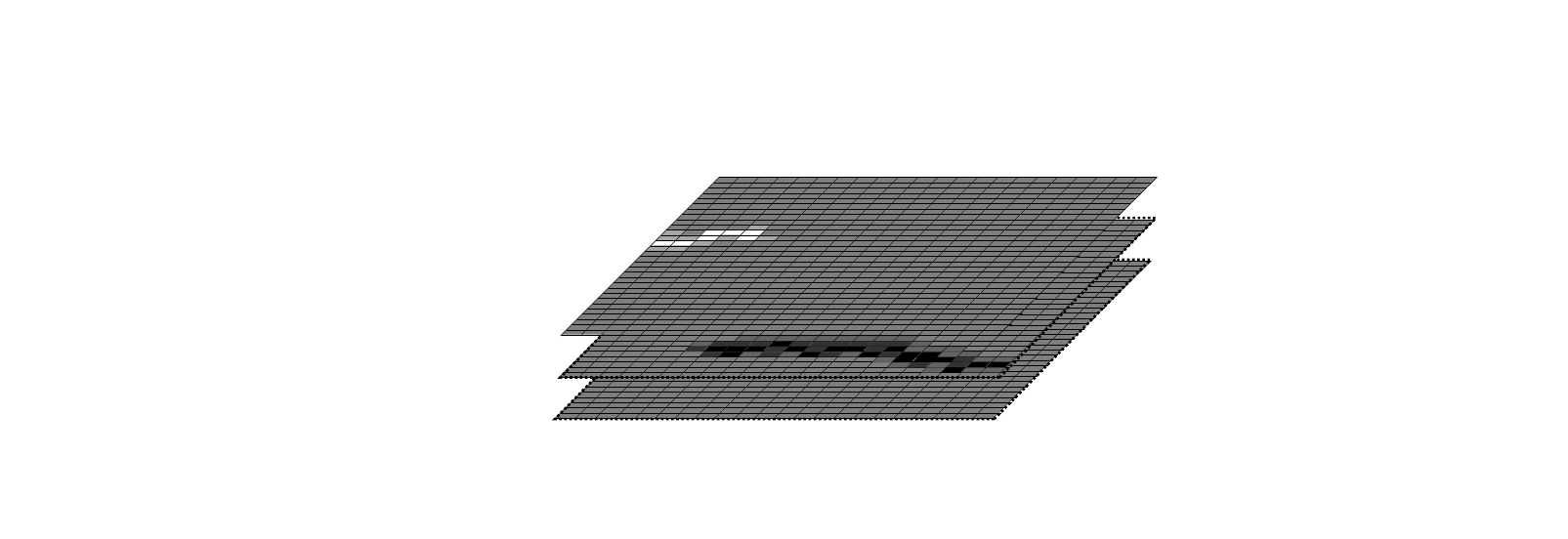}}
    \put(0.65361532,0.09276009){\color[rgb]{0,0,0}\makebox(0,0)[lt]{\begin{minipage}{0.15074685\unitlength}\raggedright $P(m_2)>0.5$\end{minipage}}}
    \put(0.32541235,0.27102143){\color[rgb]{0,0,0}\makebox(0,0)[lt]{\begin{minipage}{0.15074685\unitlength}\raggedright $P(m_1)<0.5$\end{minipage}}}
    \put(0.65711257,0.06173189){\color[rgb]{0,0,0}\makebox(0,0)[lt]{\begin{minipage}{0.15074685\unitlength}\raggedright $P(m_3)=0.5$\end{minipage}}}
    \put(0.73785319,0.14778077){\color[rgb]{0,0,0}\makebox(0,0)[lt]{\begin{minipage}{0.15074685\unitlength}\raggedright $M$\end{minipage}}}
    \put(0,0){\includegraphics[width=\unitlength,page=2]{intro.pdf}}
    \put(0.32635386,0.13971589){\color[rgb]{0,0,0}\makebox(0,0)[lt]{\begin{minipage}{0.15074685\unitlength}\raggedright $w$\end{minipage}}}
    \put(0.34086328,0.16287506){\color[rgb]{0,0,0}\makebox(0,0)[lt]{\begin{minipage}{0.15074685\unitlength}\raggedright $h$\end{minipage}}}
    \put(0.3272381,0.10953393){\color[rgb]{0,0,0}\makebox(0,0)[lt]{\begin{minipage}{0.15074685\unitlength}\raggedright $n$\end{minipage}}}
    \put(0,0){\includegraphics[width=\unitlength,page=3]{intro.pdf}}
    \put(0.02152514,0.19126339){\color[rgb]{0,0,0}\makebox(0,0)[lt]{\begin{minipage}{0.15074685\unitlength}\raggedright Local\\ Sensors\end{minipage}}}
    \put(0,0){\includegraphics[width=\unitlength,page=4]{intro.pdf}}
    \put(0.16548374,0.0603387){\color[rgb]{0,0,0}\makebox(0,0)[lt]{\begin{minipage}{0.15074685\unitlength}\raggedright Local \&\\ global\\ positioning\end{minipage}}}
    \put(0.42243827,0.30984656){\color[rgb]{0,0,0}\makebox(0,0)[lt]{\begin{minipage}{0.30241623\unitlength}\raggedright Semantical occupancy grid mapping\end{minipage}}}
    \put(0,0){\includegraphics[width=\unitlength,page=5]{intro.pdf}}
    \put(0.1719356,0.20011005){\color[rgb]{0,0,0}\makebox(0,0)[lt]{\begin{minipage}{0.15074685\unitlength}\raggedright Inverse\\ Sensor\\ Modeling\end{minipage}}}
    \put(0,0){\includegraphics[width=\unitlength,page=6]{intro.pdf}}
    \put(0.43971678,0.33377056){\color[rgb]{0,0,0}\makebox(0,0)[lt]{\begin{minipage}{0.30241623\unitlength}\raggedright Inverse sensor model interfaces\end{minipage}}}
    \put(0.87077946,0.10047677){\color[rgb]{0,0,0}\makebox(0,0)[lt]{\begin{minipage}{0.15074685\unitlength}\raggedright Local\\ applications\end{minipage}}}
    \put(0.87407199,0.25540523){\color[rgb]{0,0,0}\makebox(0,0)[lt]{\begin{minipage}{0.15074685\unitlength}\raggedright Remote\\ ISM\\ acquirers\end{minipage}}}
    \put(0.16984912,0.30410824){\color[rgb]{0,0,0}\makebox(0,0)[lt]{\begin{minipage}{0.15074685\unitlength}\raggedright Remote\\ ISM\\ aquisition\end{minipage}}}
    \put(0.05915795,0.11075588){\color[rgb]{0,0,0}\makebox(0,0)[lt]{\begin{minipage}{0.30241623\unitlength}\raggedright ISM database\end{minipage}}}
  \end{picture}
\endgroup

\end{scriptsize}
\caption{Semantic occupancy grid mapping framework}
\label{fig:framework}
\end{figure*}

While the authors have already provided a proof-of-concept of semantical grid mapping approaches in agriculture \cite{Kragh2016}, requisite information and instructions for building sensor models based on sensors and other data sources is still lacking.
In contrast to robotic and automotive approaches, where grid mapping based applications are well known, agricultural environments and applications especially vary greatly and therefore have to be treated accordingly.
With respect to \autoref{fig:framework} and \cite{Korthals2017}, this contribution focuses on the \textit{Inverse Sensor Modeling} component.

The paper is organized as follows:
\autoref{sec:RelatedWork} presents a brief introduction to occupancy grid maps, their extension to the semantical representation.
\autoref{sec:Approach} presents the gathered experience and approaches to building sensor models derived from previous agricultural research projects.
Finally, \autoref{sec:Outlook} presents further ideas and points to next steps in agricultural applications in Industry 4.0.

\section{Related Work}
\label{sec:RelatedWork}

Occupancy grid maps are used in static obstacle detection for robotic systems, which are a well-known and a commonly studied scientific field \cite{Haehnel2004, Thrun2005, stachniss09}.
They are a component of almost all navigation and collision avoidance systems designed to maneuver through cluttered environments. Another important application is the creation of obstacle maps for traversing an unknown area and the recognition of known obstacles, so supporting the localization.
Recently, occupancy grid maps have been applied to combine LiDAR and RADAR in automotive applications, with the goal of creating a harmonious, consistent and complete representation of the vehicle's environment as a basis for advanced driver assistance systems \cite{Garcia2008, Bouzouraa2010, Winner2015}.

\subsection{Occupancy Grid Mapping}

Two-dimensional occupancy grid maps (OGM) were originally introduced by Elfes \cite{Elfes1990}.
In this representation, the environment is subdivided into a regular array or a grid of quadratic cells.
The resolution of the environment representation directly depends on the size of the cells.
In addition to this compartmentalization of space, a probabilistic measure of occupancy is associated with each cell.
This measure takes any real number in the interval $[0, 1]$ and describes one of the two possible cell states: unoccupied or occupied.
An occupancy probability of $0$ represents a space that is definitely unoccupied, and a probability of $1$ represents a space that is definitely occupied.
A value of $0.5$ refers to an unknown state of occupancy.

An occupancy grid is an efficient approach to representing uncertainty, combining multiple sensor measurements at the decision level, and to incorporating different sensor models \cite{Winner2015}.
To learn an occupancy grid $M$ given sensor information $z$, different update rules exist \cite{Haehnel2004}.
For the authors' approach, a Bayesian update rule is applied to every cell $m \in M$ at position $(w,h)$ as follows:
Given the position $x_t$ of a vehicle at time $t$, let $x_{1:t}=x_1,\ldots,x_t$ be the positions of the vehicle's individual steps until $t$, and $z_{1:t}=z_1,\ldots,z_t$ the environmental perceptions.
For each cell $m$ of the occupancy probability grid the probability that this cell is occupied by an obstacle.
Thus, occupancy probability grids seek to estimate 
\begin{equation}
P\left( m | z_{1:t}, x_{1:t} \right) = \Odd^{-1} \left(\prod_{t=1}^{T}\underbrace{\frac{P\left( m | z_{t}, x_{t} \right)}{1-P\left( m | z_{t}, x_{t}\right)}}_{\Odd \left(P\left( m | z_{t}, x_{t}\right)\right)}\right) \\
\end{equation}
This equation already describes the online capable, recursive update rule that populates the current measurement $z_t$ to the grid, where $P\left( m | z_{1:t}, x_{1:t} \right)$ is the so called inverse sensor model (ISM).
The ISM is used to update the OGM in a Bayesian framework, which deduces the occupancy probability of a cell, given the sensor information.

\subsection{Extension to Agriculture Applications}

The adaptation of OGM techniques to agricultural applications appears to be merely a matter of time but is not that obvious and intuitive to apply on the second sight.
Robotic and automotive applications have in common that they both want to detect non-traversable areas or objects occupying their path.
Such unambiguous information is used to quantify the whole environment sufficiently for all derivable tasks, such as path planning or obstacle avoidance, to be completed.
When assumptions like a flat operational plane or minimum obstacle heights are made, sensors frustums oriented parallel to the ground are sufficient for all tasks

In agricultural applications, obstacle recognition is not essential as they act on and process their environment.
Therefore, quantification of the environment involves features such as processed areas, processability, crop quality, density, and maturity level in addition to traversability.
In order to map these features, single occupancy grid maps are no longer sufficient and therefore, semantic occupancy grid maps that allow different classification results to be mapped are used.
Furthermore, sensor frustums are no longer oriented parallel to the ground, but rather oriented at an angle to gather necessary crop information (cf. \autoref{fig:sensor_orientation}).

The extension to semantic occupancy grid maps (SOGM) or inference grids is straightforward and is defined by an OGM $M$ with $W$ cells in width, $H$ cells in height, and $N$ semantic layers (c.f. \autoref{fig:framework}):
\begin{equation}
M :
\left\lbrace 1, \ldots, W\right\rbrace
\times
\left\lbrace 1, \ldots, H\right\rbrace
\rightarrow
m = \left\lbrace 0, \ldots, 1\right\rbrace^N
\end{equation}
Compared to a single layer OGM which allows the classification into three classes $\left\lbrace\textrm{occupied}, \overline{\textrm{occupied}}, \textrm{unknown}\right\rbrace$, the SOGM supports a maximum of $\left|\left\lbrace\textrm{occupied}, \overline{\textrm{occupied}}, \textrm{unknown}\right\rbrace\right|^N=3^N$ different classes allowing much higher differentiability in environment and object recognition.
The corresponding ISMs are fused by means of the occupancy grid algorithm to their nth associated semantical occupancy grid.

The location of information in the maps is required to be completed by \textit{mapping under known poses} approaches \cite{Thrun2005}.
As proposed by REP-105\footnote{http://www.ros.org/reps/rep-0105.html} and realized by the authors in \cite{Korthals2017}, information is mapped locally via Kalman filtered odometry and inertial navigation measurement.
The maps themselves are globally referenced which on the one side allows smooth local mapping in the short term without the discrete jumps caused by global positioning systems using a Global Navigation Satellite System (GNSS), but also allows global consistent storing and loading of information.

While the actual features are very diverse of agriculture applications, this publication does not primarily focus on classification, but rather on geographical interpretation and sensor building.

\section{Explicit ISM Generation for Specific Sensors}
\label{sec:Approach}

\subsection{Local Sensor Based ISM}

\subsubsection{LiDAR based Mapping}

\begin{figure}
\centering
\begin{footnotesize}
\def\svgwidth{0.8\columnwidth}
\begingroup
  \makeatletter
  \providecommand\color[2][]{
    \errmessage{(Inkscape) Color is used for the text in Inkscape, but the package 'color.sty' is not loaded}
    \renewcommand\color[2][]{}
  }
  \providecommand\transparent[1]{
    \errmessage{(Inkscape) Transparency is used (non-zero) for the text in Inkscape, but the package 'transparent.sty' is not loaded}
    \renewcommand\transparent[1]{}
  }
  \providecommand\rotatebox[2]{#2}
  \ifx\svgwidth\undefined
    \setlength{\unitlength}{479.35577152bp}
    \ifx\svgscale\undefined
      \relax
    \else
      \setlength{\unitlength}{\unitlength * \real{\svgscale}}
    \fi
  \else
    \setlength{\unitlength}{\svgwidth}
  \fi
  \global\let\svgwidth\undefined
  \global\let\svgscale\undefined
  \makeatother
  \begin{picture}(1,0.52253568)
    \put(0,0){\includegraphics[width=\unitlength,page=1]{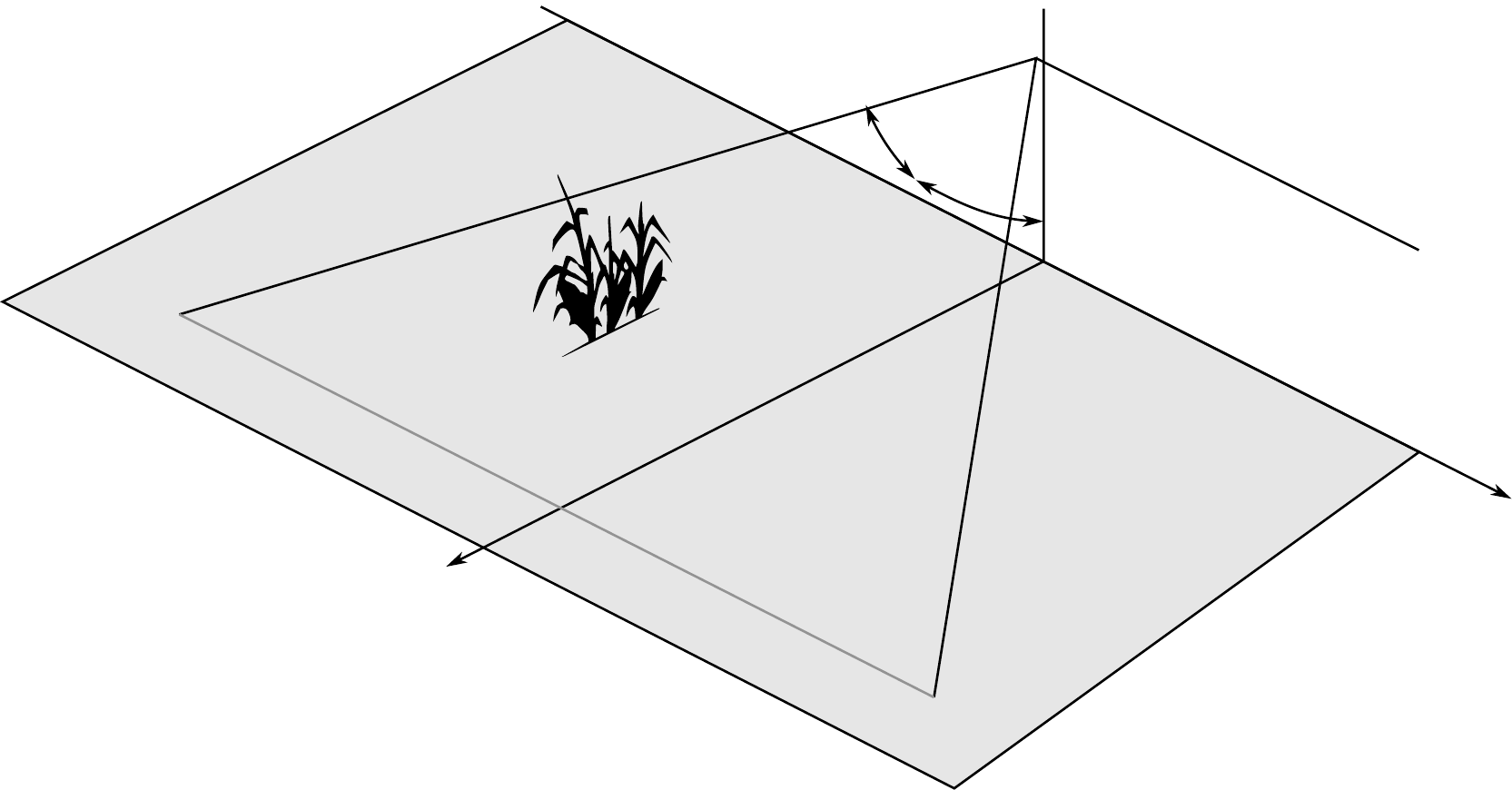}}
    \put(0.50560995,0.48929867){\color[rgb]{0,0,0}\makebox(0,0)[lb]{\smash{LiDAR}}}
    \put(0.6621679,0.50732771){\color[rgb]{0,0,0}\makebox(0,0)[lb]{\smash{Z}}}
    \put(0.90528728,0.30377654){\color[rgb]{0,0,0}\makebox(0,0)[lb]{\smash{}}}
    \put(0.8837229,0.3009788){\color[rgb]{0,0,0}\makebox(0,0)[lb]{\smash{A}}}
    \put(0.97119711,0.21446592){\color[rgb]{0,0,0}\makebox(0,0)[lb]{\smash{Y}}}
    \put(0.03930828,0.31214894){\color[rgb]{0,0,0}\rotatebox{-27.259828}{\makebox(0,0)[lb]{\smash{Ground}}}}
    \put(0,0){\includegraphics[width=\unitlength,page=2]{lidar_3.pdf}}
    \put(0.53326914,0.39154486){\color[rgb]{0,0,0}\makebox(0,0)[lb]{\smash{S}}}
    \put(0.59214182,0.42825943){\color[rgb]{0,0,0}\makebox(0,0)[lb]{\smash{$\gamma$}}}
    \put(0.64515916,0.39563332){\color[rgb]{0,0,0}\makebox(0,0)[lb]{\smash{$\phi$}}}
    \put(0,0){\includegraphics[width=\unitlength,page=3]{lidar_3.pdf}}
    \put(0.26478378,0.13189574){\color[rgb]{0,0,0}\makebox(0,0)[lb]{\smash{X}}}
    \put(0,0){\includegraphics[width=\unitlength,page=4]{lidar_3.pdf}}
    \put(0.33840818,0.25269373){\color[rgb]{0,0,0}\makebox(0,0)[lb]{\smash{$Z_S\left(X_S, Y_S\right)$}}}
  \end{picture}
\endgroup

\end{footnotesize}
\caption{Ground oriented LiDAR for crop rectification}
\label{fig:sensor_orientation}
\end{figure}

LiDAR sensors measure the distance to an object and depending on their capabilities, also the reflectance.
The distance can directly be used to deduce free (s.t. the area between the measured distance and the sensor) and occupied space (s.t. the location of measured distance) in a planar environment.
This is commonly utilized for robotic and automotive tasks, where a well-known inverse sensor modelling technique directly derives the corresponding ISM.
In agriculture, however, it is common for LiDAR sensors to face downwards as shown in \autoref{fig:sensor_orientation}, in order to detect the soil or crop that needs to be processed.
This results in the circumstance that the measurement can only be taken at the corresponding target point, and no implications can be done along the measurement.

\begin{figure}
    \centering
    \includegraphics[width=0.85\columnwidth]{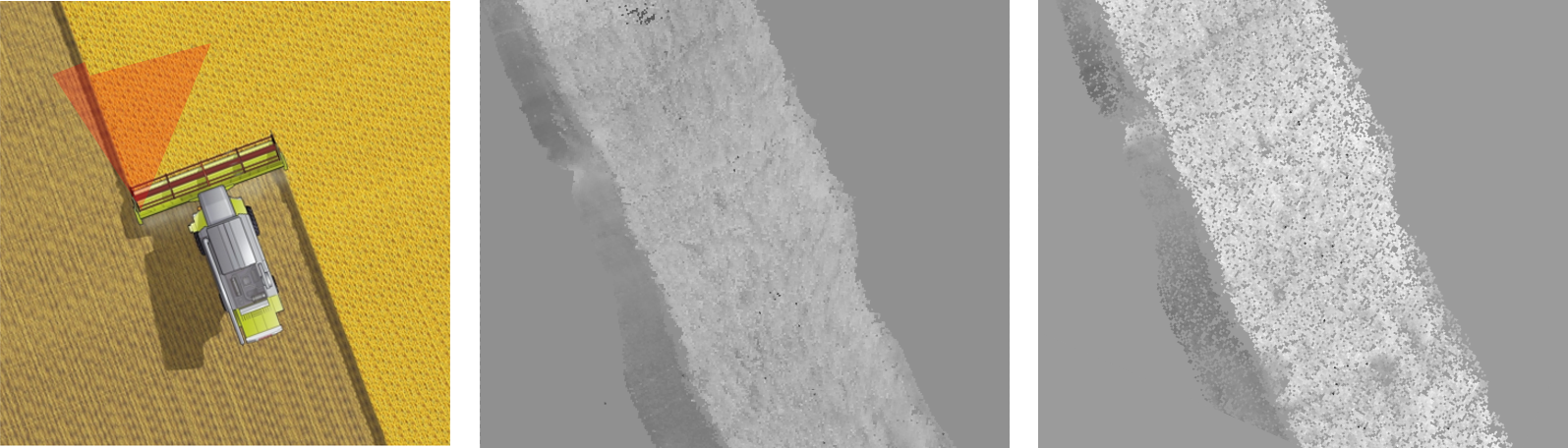}
    \caption{Harvesting scenario (left), resulting SOGM from crop classification ISM with (middle) and without (right) error propagation}
    \label{fig:lidar_example}
\end{figure}

Naively mapping the related classification in the point of measurement in the vehicles coordinate frame would result in scattered maps from which further applications are hardly derivable (c.f. \autoref{fig:lidar_example}).
Therefore, the actual Gaussian measurement uncertainty $\sigma_S$ needs to be introduced as in the common planar model, but with its appropriate error propagation.
Assuming $\sigma_\phi$, $\sigma_\xi$, $\sigma_\gamma$ beeing gaussian noise in the angular positioning caused by vehicle's steering, and $\sigma_x$, $\sigma_y$, $\sigma_z$ to be the positioning caused by vibrations of the vehicle it is possible to calculate the resulting full covariance matrix $\sum_{X_S}$ at the point of interest as follows:
First, the transformation of the scalar distance measurement $S$ in the LiDAR frame to the euclidean point $X_S$ in the vehicle frame is
\begin{equation}
X_S = \left( \begin{array}{ccc} c_\phi c_{\gamma}  \\ c_\xi s_\gamma + c_{\gamma} s_\phi s_\xi \\ s_\xi s_\gamma - c_\gamma s_\phi c_\xi\end{array}\right) S  + \operatorname{T}(x,y,z)
\end{equation}
where $\operatorname{T}$ is the translation between the sensor and the vehicle frame.
For error propagation, the functions need to be linearized by calculating the Jacobian:
\begin{equation}
\begin{split}
&J^\textrm{T} =\\
&\left( \begin{array}{ccc}
  c_\phi c_{\gamma} & c_\xi s_\gamma + c_{\gamma} s_\phi s_\xi & s_\xi s_\gamma - c_\gamma s_\phi c_\xi \\
  - S s_\phi c_{\gamma} & S c_{\gamma} c_\phi s_\xi & - S c_\gamma c_\phi c_\xi \\
  - S c_\phi s_{\gamma} & S c_\xi c_\gamma - S s_{\gamma} s_\phi s_\xi & Ss_\xi c_\gamma + S s_\gamma s_\phi c_\xi \\
  0 & - S s_\xi s_\gamma + S c_{\gamma} s_\phi c_\xi & S c_\xi s_\gamma + S c_\gamma s_\phi s_\xi    
\end{array}\right)^\textrm{T}
\end{split}
\end{equation}

\begin{equation}
\sum_{X_S} = J \operatorname{diag}(\sigma_s^2,\sigma_\phi^2, \sigma_\gamma^2, \sigma_\xi^2) J^{\text{T}} +  \operatorname{diag}(\sigma_x^2,\sigma_y^2, \sigma_z^2)
\label{eq:lidar_full_cov}
\end{equation}
The Jacobian is a function of its arguments $J(S,\phi,\gamma,\xi)$, which means that it is required to be evaluated for every new sensor measurement.
\autoref{eq:lidar_full_cov} describes the full covariance matrix which can be applied to calculate the uncertainty distribution for every measurement.

Two assumptions have been made in this model to make the error model tractable:
first, that the uncertainty in angular movements resides in the coordinate frame of the laser scanner and second, that the uncertainty in translation is uncorrelated from the angular ones.
The assumptions do not fully hold, due to the fact that rolling, pitching and yawing do not occure in the laser scanner frame, but in some other arbitrary frame, depending on the current ground conditions and vehicle's steering.
To simplify the model even more, the uncertainty in $z$ can be omitted, because in the later sensor modeling component, only the projection into the xy-plane is important.
Further, rolling is omitted as it is negligible in comparison to the other influences \cite{Konrad2012}:
\begin{equation}
\begin{split}
X'_S = \left( \begin{array}{ccc} c_\phi c_{\gamma}  \\ s_\gamma  \\ - c_\gamma s_\phi \end{array}\right) S  + \operatorname{T}(x,y,z)\\
J' = \left( \begin{array}{cccc}
c_\phi c_{\gamma} &  - S s_\phi c_{\gamma} & - S c_\phi s_{\gamma} \\
s_\gamma & 0 & S c_\gamma  \\
\end{array}\right)\\
\sum_{X'_S} = J' \operatorname{diag}(\sigma_s^2,\sigma_\phi^2, \sigma_\gamma^2) J'^{\text{T}} +  \operatorname{diag}(\sigma_x^2,\sigma_y^2)
\end{split}
\end{equation}

The influences of error propagation are depicted in \autoref{fig:lidar_example} where a two class classifier for crop derives the ISMs which are mapped to the global coordinate system.
The resulting map without error propagation is very sparse which makes further functionality derivation without heuristical post processing unfeasible.
Introducing error propagation and respecting the model uncertainties, on the other hand, results in a much more sufficient and consistent map where further classification can easily be applied.

Further improvements in classification can be achieved by first mapping the raw LiDAR data to a globally referenced representation from which further ISMs with much higher quality can be derived.
More advanced LiDAR systems scanning in multiple planes bypass the raw mapping and directly enable rich classifiers like Support Vector Machines to process the data as proposed by \cite{Kragh2016}.

\subsubsection{Inverse Perspective Mapping}

\begin{figure}
    \centering
    \includegraphics[width=0.40\textwidth]{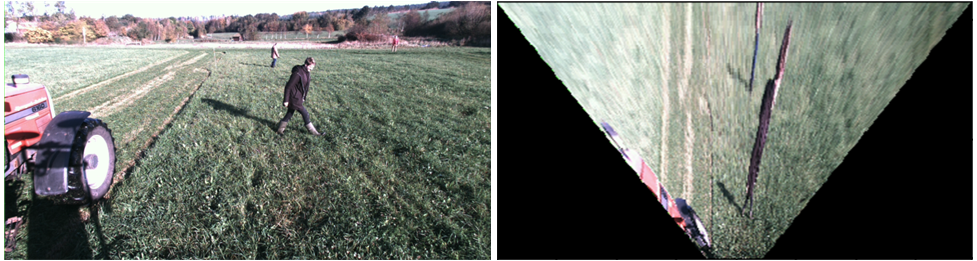}
    \caption{Inverse Perspective Mapping of RGB image}
    \label{fig:ipm_rgb}
\end{figure}

Inverse Perspective Mapping (IPM) is a geometrical transformation that projects an image to a ground plane surface as shown in \autoref{fig:ipm_rgb}.
For a flat surface, the perspective effect is removed by transforming the viewpoint from a camera view to a birds eye view.
This technique has been used in automotive applications where assumptions about camera pose and a flat world with respect to the street are sufficient \cite{Bertozzi1996, Konrad2012}.
However, even slight deviations in camera inclination and height result in large errors, more advanced, adaptive techniques have been developed which calculate the camera pose online by using the borders of the road or lane markers \cite{Simond2003}.

\begin{figure}
    \centering
    \includegraphics[width=0.40\textwidth]{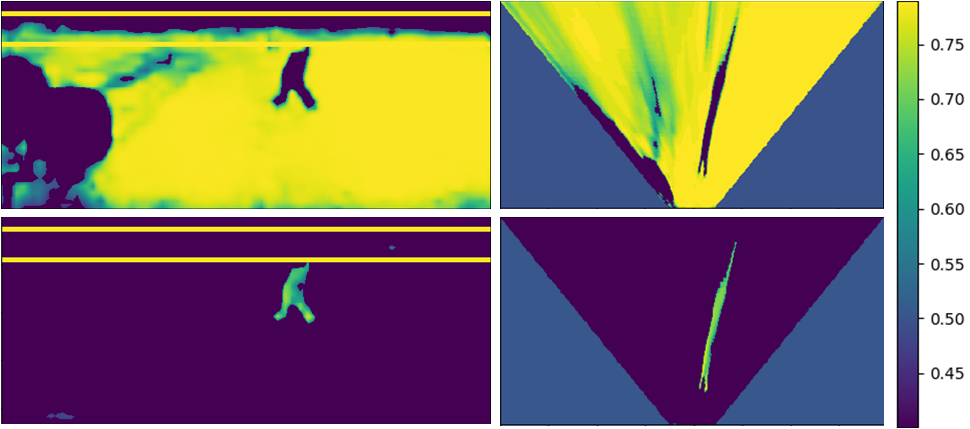}
    \caption{(Left) Grass and human predictions in a mowing application classified by a fully convolutional network for semantic segmentation \cite{Shelhamer2017} and the corresponding ISMs generated by IPM (right)}
    \label{fig:ipm_ss_fcn}
\end{figure}

However, an unstructured agricultural environments does permit such dynamic techniques and thus, they are either treated as a static scenario, where the camera pose relative to ground surface does not change, or the transformation between the extrinsic and flat plane is calculated dynamically with support of an inertial measurement unit (IMU).
The whole IPM for mapping image coordinates $\left.{\mathbf{x}_{\text{P}}}\right|_{\text{px}} = \left( u,v,1 \right)^{\text{T}}$ to surface $\left.{\mathbf{x}_{\text{FP}}}\right|_{\text{m}} = \left( x, y, z\equiv 0, 1\right)^{\text{T}}$ is defined by three parameter transformations: the intrinsic ${^{\text{P}}}\mathbf{T}_{\text{C}}$ from the camera perspective to the camera frame, the extrinsic ${^{\text{C}}}\mathbf{T}_{\text{V}}$ from the camera frame to the vehicle frame, and ${^{\text{V}}}\mathbf{T}_{\text{FP}}$ which transforms from the vehicle frame to the flat plane (FP) frame.
This leads to 
\begin{equation}
\left.{\mathbf{x}_{\text{P}}}\right|_{\text{px}} = {^{\text{P}}}\mathbf{T}_{\text{C}} \cdot {^{\text{C}}}\mathbf{T}_{\text{V}} \cdot {^{\text{V}}}\mathbf{T}_{\text{FP}} \cdot \left.{\mathbf{x}_{\text{FP}}}\right|_{\text{m}}
\label{eq:IPM}
\end{equation}
To build the actual ISM, the image first needs to be classified and then transformed to the flat plane by means of \autoref{eq:IPM} (c.f. \autoref{fig:ipm_ss_fcn}).

Values of an ISM are the probability of a grid cell being occupied by a giving classification.
As indicated in \autoref{fig:ipm_ss_fcn}, the area that is not visible by the camera is set to 0.5 to represent the fact no information is provided for areas that are not visible to the camera.
Visible areas with no detections are set below 0.5 to indicate that the area is not expected to be occupied by the given class.
Values above 0.5 indicate that the area is expected to be occupied by the given class.

For detecting flat class elements such as road-lane markings or grass, the IPM algorithm is able to provide good approximations of the actual inverse perspective mapping.
Elevated elements violate the IPM ground plane assumption and will stretch elements unnaturally and incorrectly across large areas as indicated in \autoref{fig:ipm_rgb}.

\begin{figure}
    \centering
    \begin{minipage}{0.15\textwidth} 
    \includegraphics[width=\textwidth]{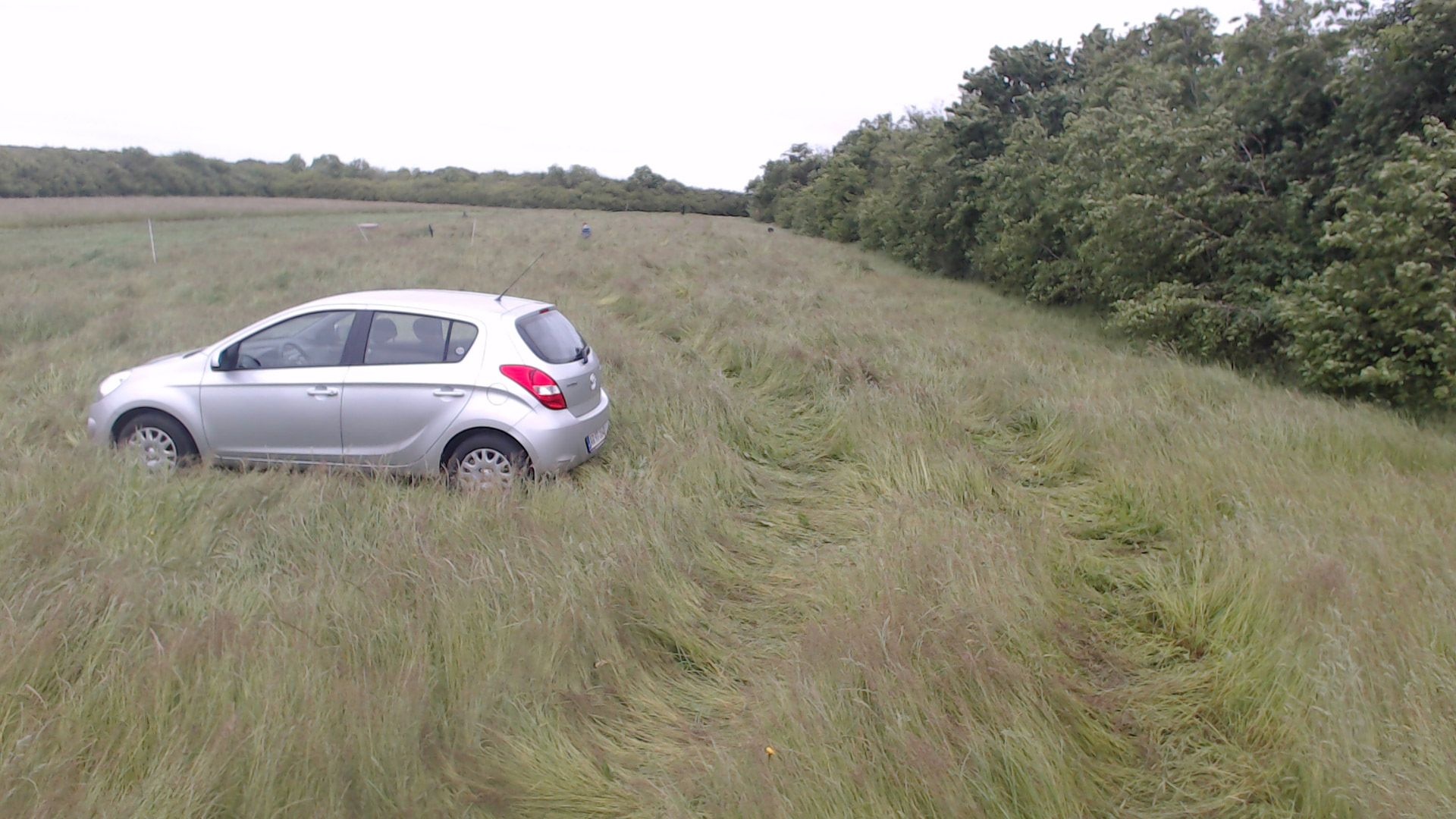}
	\end{minipage}
	\hfill
    \begin{minipage}{0.15\textwidth} 
    \includegraphics[width=\textwidth]{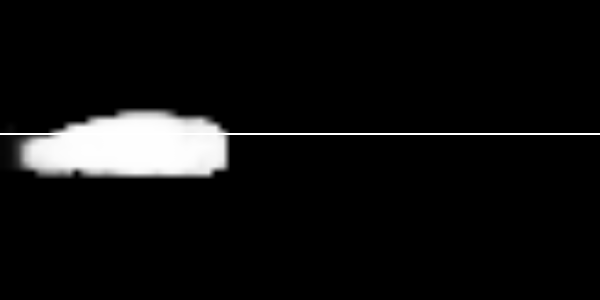}
	\end{minipage}
	\hfill
	\begin{minipage}{0.15\textwidth}
	\includegraphics[width=\textwidth]{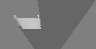}
	\end{minipage}
    \caption{Input image (left), classification based on semantic segmentation (middle) and corresponding ISM with detection cut-off after class occurrence along the focal axis}
    \label{fig:IPM_naive}
\end{figure}

To avoid the stretching artifacts of tall objects, different approaches are proposed.
A naive approach for pixel based classifiers states that all objects classified as being other than ground are standing perpendicular on the ground.
Therefore, one can perform a ray trace along the focal axis and mark all cells behind a detected object as unknown (c.f. \autoref{fig:IPM_naive}) \cite{Kohlbrecher2011, Kragh2016}.

Another approach generates three dimensional object location hypotheses by first estimating the distance to the corresponding detection.
This can be achieved by either using the abovementioned naive approach or using a depth sensor like a stereo camera or LiDAR which is registered to the camera.

\begin{figure}
    \centering
    \includegraphics[width=0.45\textwidth]{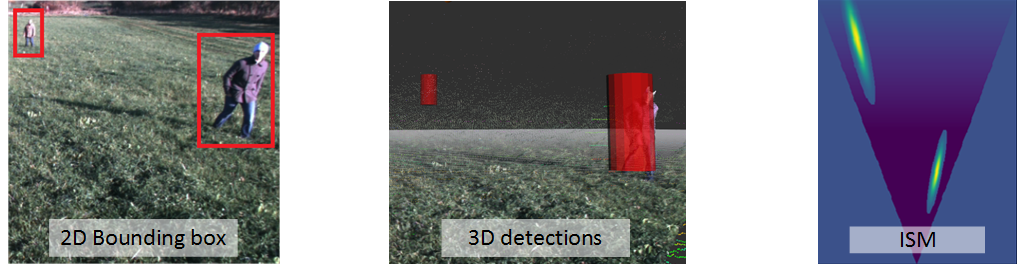}
    \caption{Bounding box detection to ISM}
    \label{fig:ipm_bb}
\end{figure}

Second, when using classifiers like YOLO \cite{Redmon2016} which offers classified bounding boxes, the four bounding box corners are mapped to real world coordinates using the estimated distance to a detection and the intrinsic camera parameters.
The bounding box position and extent are derived in 3D and is represented as depicted in \autoref{fig:ipm_bb} by cylinder specified by a center, height, and width.

Detections are mapped to values above 0.5 with a Gaussian distribution to indicate the existence of an obstacle with corresponding localization uncertainties.
The localization uncertainty for the camera depends on the radial coordinate (distance to the object) and angular coordinate (angle to object), where accuracy degrades with increasing distance and angle.
The procedure for converting a 2D bounding box to an ISM using distance estimates is presented in \autoref{fig:ipm_bb}. 
Using the estimated distance of a detected object and the intrinsic camera parameters, the four bounding box corners are mapped to world coordinates.

\begin{figure}
\centering
\begin{footnotesize}
\def\svgwidth{0.85\columnwidth}
\begingroup
  \makeatletter
  \providecommand\color[2][]{
    \errmessage{(Inkscape) Color is used for the text in Inkscape, but the package 'color.sty' is not loaded}
    \renewcommand\color[2][]{}
  }
  \providecommand\transparent[1]{
    \errmessage{(Inkscape) Transparency is used (non-zero) for the text in Inkscape, but the package 'transparent.sty' is not loaded}
    \renewcommand\transparent[1]{}
  }
  \providecommand\rotatebox[2]{#2}
  \ifx\svgwidth\undefined
    \setlength{\unitlength}{275.03746934bp}
    \ifx\svgscale\undefined
      \relax
    \else
      \setlength{\unitlength}{\unitlength * \real{\svgscale}}
    \fi
  \else
    \setlength{\unitlength}{\svgwidth}
  \fi
  \global\let\svgwidth\undefined
  \global\let\svgscale\undefined
  \makeatother
  \begin{picture}(1,0.30697908)
    \put(0,0){\includegraphics[width=\unitlength,page=1]{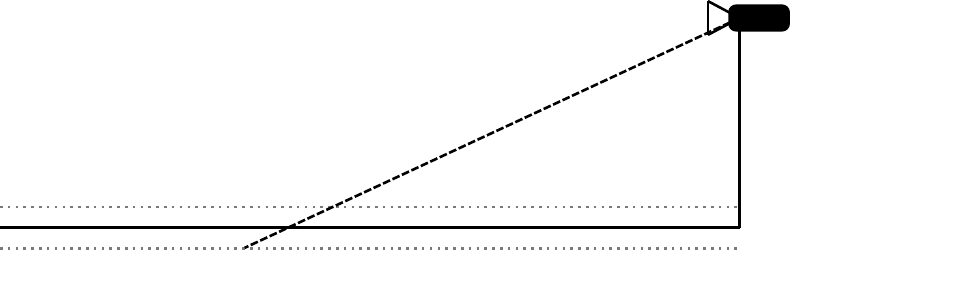}}
    \put(-0.00750104,0.37361525){\color[rgb]{0,0,0}\makebox(0,0)[lt]{\begin{minipage}{0.42386777\unitlength}\raggedright \end{minipage}}}
    \put(-0.12877428,0.47958221){\color[rgb]{0,0,0}\makebox(0,0)[lt]{\begin{minipage}{0.35086829\unitlength}\raggedright \end{minipage}}}
    \put(0.29899266,0.19347144){\color[rgb]{0,0,0}\makebox(0,0)[lb]{\smash{focal mapping}}}
    \put(0.34454475,0.62793592){\color[rgb]{0,0,0}\makebox(0,0)[lt]{\begin{minipage}{0.42504523\unitlength}\raggedright \end{minipage}}}
    \put(-0.01927511,0.61851664){\color[rgb]{0,0,0}\makebox(0,0)[lt]{\begin{minipage}{0.51688316\unitlength}\raggedright \end{minipage}}}
    \put(-0.13348394,0.59732325){\color[rgb]{0,0,0}\makebox(0,0)[lt]{\begin{minipage}{0.42975485\unitlength}\raggedright \end{minipage}}}
    \put(0,0){\includegraphics[width=\unitlength,page=2]{flat_plane.pdf}}
    \put(0.04515747,0.07893476){\color[rgb]{0,0,0}\makebox(0,0)[lb]{\smash{flat plane}}}
    \put(0.7866558,0.16530282){\color[rgb]{0,0,0}\makebox(0,0)[lb]{\smash{$h$}}}
    \put(0.52882105,0.07221369){\color[rgb]{0,0,0}\makebox(0,0)[lb]{\smash{$d$}}}
    \put(0,0){\includegraphics[width=\unitlength,page=3]{flat_plane.pdf}}
    \put(0.85595806,0.27755431){\color[rgb]{0,0,0}\makebox(0,0)[lb]{\smash{camera}}}
    \put(0.82369237,0.06293025){\color[rgb]{0,0,0}\makebox(0,0)[lb]{\smash{$\sigma_h$}}}
    \put(0.37348405,0.00765166){\color[rgb]{0,0,0}\makebox(0,0)[lb]{\smash{$\sigma_d$}}}
    \put(0,0){\includegraphics[width=\unitlength,page=4]{flat_plane.pdf}}
  \end{picture}
\endgroup

\end{footnotesize}
\caption{Simplified error assumption in flat plane assumption according to height}
\label{fig:flat_plane}
\end{figure}

\begin{figure}
    \centering
    \includegraphics[width=0.7\columnwidth]{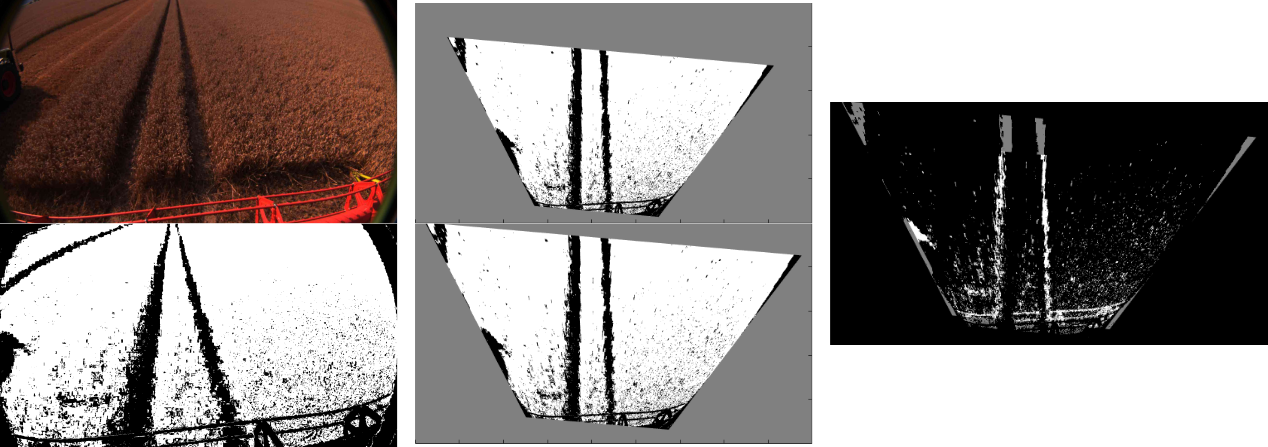}
    \caption{RGB input image and scanline based classification for crop plane (left), inverse perspective mapping of classification for crop and ground plane (middle) and corresponding fused contradicting ISM (right)}
    \label{fig:ipm_contradicting}
\end{figure}

Lastly, the concept of contradicting IPM is introduced for crop processing in harvesting scenarios.
In comparison with the abovementioned IPM scenarios, this discrimination is necessary as the camera rectifies no common ground in the lower areas of the image as depicted in \autoref{fig:ipm_contradicting} which refutes former assumptions.
Neglecting this fact would result in drastically wrong localization of detections, as visualized in \autoref{fig:flat_plane}, which indicates that the localization error $\sigma_d$ in depth $d$ depends on the error $\sigma_h$ of height $h$ as follows:
\begin{equation}
\sigma_d = \frac{d}{h} \sigma_h\textrm{.}
\end{equation}
If this simple error propagation is applied to a hypothetical example of small crop with for example a height of 0.5 meters and a camera installation height of 1.5 meters where a feature 10 meters away should be mapped, the resultant error is one of 3 meters.
Therefore, two flat plane assumptions are calculated, one for the ground and one for the crop height resulting in two different ISMs.
These can then be combined by Dempster’s rule of combination leading to contradictions \cite{Shafer1976}, which is visualized in \autoref{fig:ipm_contradicting}.
From the emerging contradictions in \autoref{fig:ipm_contradicting} (right), it can be seen that vehicle traces appear which are actually the contradicting occlusion in both IPMs.

\subsubsection{Ambiguous Sensor Mapping}

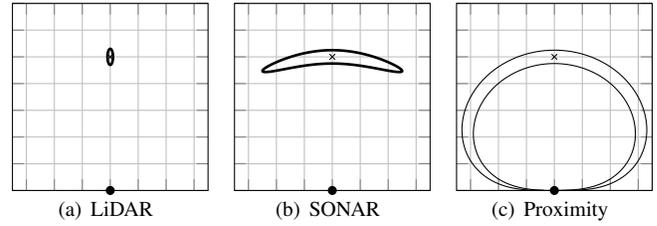
\begin{figure}
\centering
\setlength\figurewidth{0.195\textwidth}
\setlength\figureheight{0.14\textwidth}
\tiny
\subfigure[LiDAR]{\input{sensorConeLidar.tex}\label{fig:sensorConeLidar}}\hfill
\subfigure[SONAR]{\input{sensorConeSonar.tex}\label{fig:sensorConeSonar}}\hfill
\subfigure[Proximity]{\input{sensorConeProx.tex}\label{fig:sensorConeProx}}\hfill
\caption{Standard error contour of qualitative sensor cones ($\Bigcdot$: Sensor position, x: Obstacle, -)}
\label{fig:cones}
\end{figure}

Ambiguous sensor readings originate from sensors with very bad angular or distance resolution by definition of the authors.
As depicted in \autoref{fig:cones} LiDAR systems can achieve very accurate positioning and are therefore the preferred sensors for mapping.
However, they are by far the most cost- and power intensive systems.
Other sensing techniques are more cost and power efficient but are commonly neglected due to their high noise or inaccuracy.
Nevertheless, the authors have demonstrated that even with poorly embedded sensors, sufficient environment detection can be achieved \cite{Korthals20161} by designing an inverse particle filter which samples from the sensors uncertainty distribution.
At present, this technique has only been applied in laboratory conditions and therefore, real agricultural applications remain pending.

\subsection{Application Models}

\begin{figure}
    \centering
	\begin{minipage}{0.2\textwidth}
	\includegraphics[width=\textwidth]{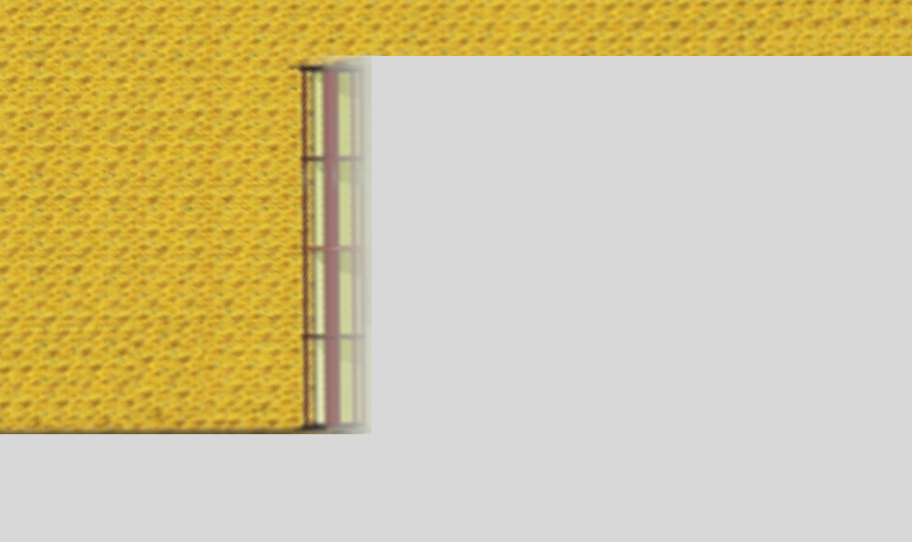}
	\end{minipage}
    \caption{Top view of crop field with an applied inverse sensor model for the cutter bar: gray shaded area being of high probability that the cutter bar has been applied on that region}
    \label{fig:ism_application}
\end{figure}

Application models are straight forward to implement and only depends on the localizing accuracy.
Building such a model is only dependent on the geometrical shape of the agricultural implement.
That means on the other hand, that ISM is a static and primitive shape in the local frame of the vehicle which leaves a probabilistic footprint where the implement has been applied to the crop as depicted in \autoref{fig:ism_application}.
When incorporating inaccurate localization, the shape needs to be transformed accordingly.

\subsection{Map Services}

\begin{figure}
\centering
\begin{scriptsize}
\def\svgwidth{0.75\columnwidth}
\begingroup
  \makeatletter
  \providecommand\color[2][]{
    \errmessage{(Inkscape) Color is used for the text in Inkscape, but the package 'color.sty' is not loaded}
    \renewcommand\color[2][]{}
  }
  \providecommand\transparent[1]{
    \errmessage{(Inkscape) Transparency is used (non-zero) for the text in Inkscape, but the package 'transparent.sty' is not loaded}
    \renewcommand\transparent[1]{}
  }
  \providecommand\rotatebox[2]{#2}
  \ifx\svgwidth\undefined
    \setlength{\unitlength}{478.74014787bp}
    \ifx\svgscale\undefined
      \relax
    \else
      \setlength{\unitlength}{\unitlength * \real{\svgscale}}
    \fi
  \else
    \setlength{\unitlength}{\svgwidth}
  \fi
  \global\let\svgwidth\undefined
  \global\let\svgscale\undefined
  \makeatother
  \begin{picture}(1,0.67497711)
    \put(0,0){\includegraphics[width=\unitlength,page=1]{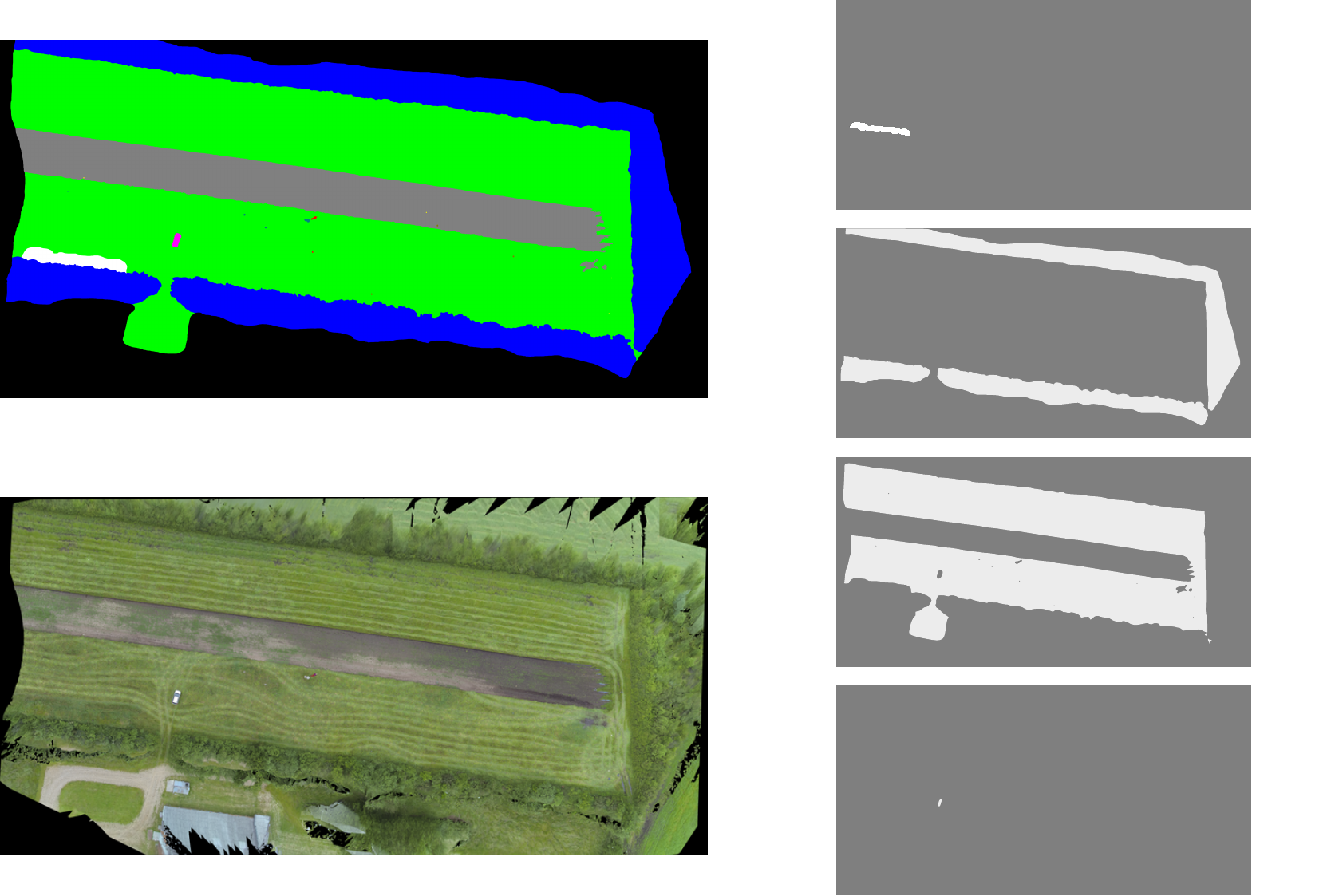}}
    \put(1.0095952,0.65320651){\color[rgb]{0,0,0}\makebox(0,0)[lt]{\begin{minipage}{0.35228123\unitlength}\raggedright \end{minipage}}}
    \put(0.96012592,0.28940573){\color[rgb]{0,0,0}\rotatebox{-90}{\makebox(0,0)[lb]{\smash{Ground}}}}
    \put(0.96012592,0.12487914){\color[rgb]{0,0,0}\rotatebox{-90}{\makebox(0,0)[lb]{\smash{Obstacle}}}}
    \put(0.96012592,0.48155668){\color[rgb]{0,0,0}\rotatebox{-90}{\makebox(0,0)[lb]{\smash{Shelterbelt}}}}
    \put(0.98477243,0.63031028){\color[rgb]{0,0,0}\rotatebox{-90}{\makebox(0,0)[lb]{\smash{Non-}}}}
    \put(0.96012592,0.66768051){\color[rgb]{0,0,0}\rotatebox{-90}{\makebox(0,0)[lb]{\smash{Processable}}}}
    \put(0,0){\includegraphics[width=\unitlength,page=2]{drone_ism.pdf}}
    \put(0.29508764,0.33089805){\color[rgb]{0,0,0}\makebox(0,0)[lb]{\smash{Classification}}}
  \end{picture}
\endgroup

\end{scriptsize}
\caption{Classification decomposition of hand labeled orthographic photograph \cite{Kragh2016}}
\label{fig:ism_geo}
\end{figure}

Geodata acquired by satellites, drones, or planes with high recording frequencies as well as its partially free availability, make this information increasingly attractive for agriculture.
In this context worth mentioning are the Sentinel program\footnote{\url{http://www.esa.int/Our_Activities/Observing_the_Earth/Copernicus/Overview4/}}, the hyperspectral system EnMap\footnote{\url{http://www.enmap.org/}}, the RapidEye constellation\footnote{\url{http://blackbridge.com/rapideye/}} as well as the start-up companies Skybox Imaging\footnote{\url{http://www.skyboximaging.com/}} and Planet Labs\footnote{\url{https://www.planet.com/}}.
In addition, the release of the long-standing Landsat archive now offers many opportunities for agricultural applications, such as the generation of profit potential maps.
There is a trend towards direct access to such data and towards appropriate image excerpts using web servers or APIs.
As part of spatial data infrastructures, data (e.g. land and terrain data) are published interoperably and often free of charge via web services.
In particular, Annex III of the INSPIRE Directive\footnote{\url{http://inspire.ec.europa.eu/}} requires EU member states to provide data.
However, for a precision farming service or a precision farming application further different data sources have to be linked (for example, weather data play a crucial role in most agricultural processes), or complex procedures and algorithms are required to derive the desired information from the data.
Subsequent downstream services will continue to play an increasingly important role in agriculture. The European Union, for example, specifically supports the development of such services based on Copernicus data by SMEs.
At the endpoint of the downstream services, information products (such as humidity maps, biomass maps and yield forecast maps) are often available, which can be integrated into other applications or devices.
The combination and the inclusion of all the information sources and their derivation for the identification of machine parameters is one essential part which can be handled by ISMs.
As an example, a static and classified drone image can be easily transferred to a semantic ISM by decomposing all classes and loading the appropriate area during operation (c.f. \autoref{fig:ism_geo}).
\section{Conclusion and Outlook}
\label{sec:Outlook}
The authors have presented an information representation as semantic grids which can be maintained among different modalities and sources.
It utilizes the idea of the ISOBUS standard, which was designed with machinery interoperability in mind, and allows every sensory source to publish or access its information in a general grid format.
The main aspect of this contribution focused on different techniques, originating from literature, practical experiments, and experience, of actually building these representations.

As the acquisition and localization of data are sufficiently solved, further research will concentrate on planning and control of such diverse data.
Furthermore, learning approaches have not been confronted in this application which directly maps a sensor reading to the appropriate locality and probability.
These techniques were introduced by Thrun \cite{Thrun2005} and have been applied by the authors.
However, following the engineering path of building inverse sensor models is far more robust and intuitive. 
At present, only a few approaches are known to the authors and therefore, more applications extending from direct control architectures up to holistic farm management systems are of great interest. 
Approaching rich control architectures in agricultural environments allows an interesting area of overlap between robotics and Industry 4.0 to emerge, s.t. simultaneously planning and processing.
Mathematical frameworks exist, where in agriculture the particular issue will driven by the information representation and how it is incorporated into environmental processing.

\subsection{Dynamic Scenario}
\label{seq:eval_dynamic}

\addtolength{\textheight}{-12cm}   
\section*{ACKNOWLEDGMENT}
This research was supported by the Cluster of Excellence Cognitive Interaction Technology 'CITEC' (EXC 277) at Bielefeld University, which is funded by the German Research Foundation (DFG) and by the German Federal Ministry of Education and Research (BMBF) within the Leading-Edge Cluster {\elqq}Intelligent Technical Systems OstWestfalenLippe{\erqq} (it's OWL) and managed by the Project Management Agency Karlsruhe (PTKA).

\bibliographystyle{IEEEtran}
\bibliography{root}

\end{document}

%% file: sensorConeLidar.tex
%
%
\begin{tikzpicture}

\begin{axis}[%
width=0.75\figurewidth,
height=\figureheight,
at={(0\figurewidth,0\figureheight)},
scale only axis,
xmin=-0.7,
xmax=0.7,
xtick={-0.6,-0.4,-0.2,0,0.2,0.4,0.6},
xticklabels={\empty},
xmajorgrids,
ymin=0,
ymax=1.4,
ytick={0,0.2,0.4,0.6,0.8,1,1.2,1.4},
yticklabels={\empty},
ymajorgrids,
axis background/.style={fill=white}
]
\addplot [color=black,solid,line width=1.0pt,forget plot]
  table[row sep=crcr]{%
-0.02	1\\
-0.0199990000083333	0.99940000999995\\
-0.0199960001333316	0.9988000799984\\
-0.0199910006749798	0.99820026998785\\
-0.0199840021332196	0.997600639948802\\
-0.0199750052078993	0.997001249843759\\
-0.0199640107987041	0.996402159611233\\
-0.0199510200050656	0.995803429159748\\
-0.0199360341260524	0.99520511836185\\
-0.0199190546602399	0.994607287048119\\
-0.0199000833055605	0.99400999500119\\
-0.0198791219591339	0.99341330194977\\
-0.0198561727170773	0.992817267562665\\
-0.0198312378742958	0.992221951442818\\
-0.0198043199242527	0.991627413121346\\
-0.0197754215587208	0.991033712051584\\
-0.0197445456675125	0.990440907603145\\
-0.0197116953381912	0.98984905905598\\
-0.0196768738557624	0.989258225594451\\
-0.0196400847023454	0.98866846630141\\
-0.0196013315568248	0.988079840152296\\
-0.019560618294483	0.987492406009234\\
-0.0195179489866121	0.986906222615148\\
-0.0194733279001075	0.986321348587889\\
-0.0194267594970406	0.985737842414372\\
-0.0193782484342129	0.985155762444729\\
-0.0193277995626903	0.984575166886471\\
-0.0192754179273178	0.98399611379867\\
-0.0192211087662154	0.983418661086153\\
-0.0191648775102539	0.98284286649371\\
-0.0191067297825121	0.98226878760032\\
-0.0190466713977143	0.981696481813393\\
-0.0189847083616488	0.981126006363033\\
-0.0189208468705677	0.980557418296308\\
-0.0188550933105669	0.979990774471551\\
-0.0187874542569476	0.979426131552673\\
-0.0187179364735587	0.978863546003495\\
-0.0186465469121207	0.978303074082102\\
-0.0185732927115302	0.977744771835221\\
-0.0184981811971463	0.97718869509261\\
-0.0184212198800577	0.976634899461481\\
-0.0183424164563321	0.976083440320935\\
-0.0182617788062462	0.975534372816426\\
-0.0181793149934977	0.974987751854247\\
-0.0180950332643993	0.97444363209604\\
-0.0180089420470535	0.973902067953326\\
-0.0179210499505105	0.973363113582069\\
-0.0178313657639066	0.972826822877256\\
-0.0177398984555857	0.972293249467511\\
-0.0176466571722024	0.971762446709731\\
-0.0175516512378075	0.971234467683748\\
-0.017454890152915	0.970709365187026\\
-0.017356383593553	0.970187191729376\\
-0.0172561414102952	0.969667999527709\\
-0.0171541736272765	0.969151840500813\\
-0.0170504904411901	0.96863876626416\\
-0.0169451022202683	0.968128828124747\\
-0.0168380195032454	0.967622077075962\\
-0.0167292529983037	0.967118563792488\\
-0.0166188135820033	0.966618338625233\\
-0.0165067122981936	0.966121451596298\\
-0.0163929603569096	0.965627952393971\\
-0.0162775691332507	0.965137890367762\\
-0.016160550166243	0.964651314523464\\
-0.0160419151576858	0.964168273518257\\
-0.0159216759709811	0.963688815655838\\
-0.0157998446299473	0.963212988881594\\
-0.015676433317617	0.962740840777806\\
-0.0155514543750186	0.962272418558892\\
-0.0154249202999421	0.961807769066682\\
-0.0152968437456898	0.961346938765739\\
-0.0151672375198102	0.960889973738708\\
-0.0150361145828179	0.960436919681712\\
-0.0149034880468974	0.959987821899778\\
-0.0147693711745918	0.959542725302311\\
-0.0146337773774764	0.9591016743986\\
-0.0144967202148181	0.958664713293367\\
-0.0143582133922189	0.958231885682359\\
-0.0142182707602455	0.957803234847975\\
-0.0140769063130447	0.957378803654944\\
-0.0139341341869433	0.956958634546029\\
-0.0137899686590349	0.956542769537791\\
-0.0136444241457523	0.956131250216386\\
-0.0134975152014253	0.955724117733402\\
-0.0133492565168262	0.955321412801748\\
-0.0131996629176996	0.954923175691582\\
-0.013048749363281	0.954529446226283\\
-0.0128965309448	0.95414026377847\\
-0.0127430228839716	0.953755667266062\\
-0.0125882405314739	0.953375695148391\\
-0.0124321993654133	0.953000385422351\\
-0.0122749149897762	0.952629775618603\\
-0.0121164031328693	0.952263902797818\\
-0.011956679645746	0.951902803546973\\
-0.011795760500622	0.951546513975693\\
-0.0116336617892777	0.951195069712638\\
-0.0114703997214491	0.95084850590194\\
-0.0113059906232071	0.950506857199693\\
-0.0111404509353243	0.950170157770482\\
-0.0109737972116317	0.949838441283969\\
-0.0108060461173628	0.949511740911526\\
-0.0106372144274871	0.949190089322919\\
-0.010467319025033	0.948873518683038\\
-0.0102963768993991	0.948562060648684\\
-0.0101244051446556	0.9482557463654\\
-0.00995142095783454	0.947954606464359\\
-0.00977744163721055	0.947658671059301\\
-0.00960248458057067	0.947367969743519\\
-0.0094265672834748	0.947082531586903\\
-0.00924970733750602	0.946802385133031\\
-0.00907192242851154	0.946527558396314\\
-0.00889323033483413	0.946258078859197\\
-0.00871364892553424	0.94599397346941\\
-0.00853319615860315	0.94573526863727\\
-0.00835189007916716	0.945481990233047\\
-0.00816974881768314	0.945234163584369\\
-0.00798679058812546	0.944991813473694\\
-0.0078030336861646	0.944754964135832\\
-0.00761849648733763	0.944523639255519\\
-0.00743319744521066	0.94429786196505\\
-0.00724715508953347	0.944077654841966\\
-0.0070603880243866	0.943863039906797\\
-0.00687291492632094	0.943654038620856\\
-0.00668475454249005	0.943450671884098\\
-0.00649592568877552	0.943252960033028\\
-0.00630644724790537	0.943060922838665\\
-0.00611633816756578	0.942874579504569\\
-0.00592561745850637	0.942693948664919\\
-0.00573430419263911	0.942519048382647\\
-0.00554241750113115	0.942349896147636\\
-0.00534997657249175	0.942186508874968\\
-0.00515700065065339	0.942028902903236\\
-0.00496350903304746	0.941877093992904\\
-0.00476952106867446	0.941731097324737\\
-0.00457505615616919	0.941590927498281\\
-0.00438013374186083	0.9414565985304\\
-0.00418477331782838	0.941328123853881\\
-0.00398899441995145	0.941205516316083\\
-0.00379281662595668	0.941088788177658\\
-0.00359625955345998	0.940977951111323\\
-0.00339934285800481	0.940873016200692\\
-0.00320208623109662	0.940773993939169\\
-0.00300450939823371	0.940680894228897\\
-0.00280663211693473	0.940593726379769\\
-0.00260847417476291	0.940512499108499\\
-0.00241005538734733	0.940437220537745\\
-0.00221139559640139	0.940367898195301\\
-0.00201251466773863	0.940304539013345\\
-0.00181343248928619	0.940247149327742\\
-0.00161416896909601	0.940195734877411\\
-0.00141474403335406	0.940150300803757\\
-0.00121517762438772	0.940110851650147\\
-0.00101548969867158	0.940077391361465\\
-0.000815700224831818	0.940049923283711\\
-0.00061582918164932	0.940028450163671\\
-0.000415896556061846	0.940012974148639\\
-0.000215922341165345	0.940003496786202\\
-1.59265342146628e-05	0.94000001902409\\
0.000184070865376169	0.940002541210076\\
0.000384049858033855	0.940011063091943\\
0.000583990446025779	0.94002558381751\\
0.000783872635459757	0.940046101934717\\
0.000983676438283413	0.940072615391769\\
0.00118338187428297	0.940105121537342\\
0.00138296897308125	0.94014361712085\\
0.00158241777613468	0.940188098292765\\
0.00178170833872919	0.940238560605008\\
0.00198082073197456	0.940294999011389\\
0.00217973504479742	0.940357407868115\\
0.00237843138593225	0.940425780934353\\
0.0025768898859105	0.940500111372852\\
0.00277509069904755	0.940580391750631\\
0.00297301400542728	0.940666614039718\\
0.00317064001288396	0.940758769617957\\
0.00336794895898154	0.940856849269867\\
0.00356492111298984	0.940960843187564\\
0.0037615367778576	0.941070740971743\\
0.00395777629218218	0.941186531632717\\
0.00415362003217568	0.941308203591517\\
0.0043490484136273	0.941435744681049\\
0.00454404189386174	0.941569142147308\\
0.0047385809736935	0.941708382650661\\
0.00493264619937668	0.941853452267173\\
0.00512621816455046	0.942004336490002\\
0.00531927751217961	0.942161020230853\\
0.00551180493649026	0.942323487821482\\
0.00570378118490042	0.942491723015265\\
0.00589518705994522	0.942665708988823\\
0.00608600342119667	0.942845428343704\\
0.00627621118717765	0.943030863108122\\
0.00646579133727007	0.943221994738755\\
0.00665472491361691	0.943418804122599\\
0.00684299302301797	0.943621271578881\\
0.0070305768388192	0.943829376861022\\
0.00721745760279534	0.944043099158668\\
0.00740361662702574	0.944262417099768\\
0.00758903529576309	0.94448730875271\\
0.00777369506729504	0.944717751628516\\
0.00795757747579832	0.944953722683091\\
0.00814066413318531	0.945195198319526\\
0.00832293673094285	0.945442154390459\\
0.00850437704196305	0.945694566200488\\
0.00868496692236601	0.945952408508642\\
0.00886468831331419	0.946215655530906\\
0.00904352324281824	0.946484280942797\\
0.00922145382753426	0.946758257881998\\
0.00939846227455205	0.947037558951042\\
0.0095745308831744	0.947322156220055\\
0.00974964204668719	0.947612021229549\\
0.00992377825411998	0.947907124993263\\
0.0100969220919972	0.948207438001068\\
0.0102690562460792	0.948512930221912\\
0.0104401635030941	0.948823571106827\\
0.0106102267524589	0.949139329591986\\
0.0107792289879902	0.949460174101801\\
0.0109471533096054	0.94978607255209\\
0.0111139829250123	0.950116992353279\\
0.0112797011513882	0.950452900413663\\
0.0114442914170487	0.950793763142718\\
0.0116077372631044	0.951139546454452\\
0.0117700223451069	0.951490215770825\\
0.0119311304346832	0.951845736025195\\
0.0120910454211586	0.952206071665835\\
0.0122497513131677	0.95257118665948\\
0.0124072322402536	0.952941044494935\\
0.0125634724544548	0.953315608186725\\
0.0127184563318801	0.953694840278792\\
0.0128721683742708	0.954078702848245\\
0.0130245932105509	0.954467157509145\\
0.0131757155983638	0.95486016541635\\
0.0133255204255965	0.955257687269397\\
0.0134739927118912	0.955659683316432\\
0.0136211176101431	0.956066113358185\\
0.0137668804079848	0.956476936751991\\
0.013911266529258	0.956892112415852\\
0.0140542615354711	0.957311598832549\\
0.0141958511272424	0.957735354053789\\
0.0143360211457309	0.958163335704404\\
0.0144747575740514	0.958595500986584\\
0.0146120465386767	0.959031806684162\\
0.0147478743108249	0.959472209166931\\
0.0148822273078319	0.959916664395011\\
0.0150150920945098	0.960365127923248\\
0.0151464553844909	0.960817554905664\\
0.0152763040415555	0.961273900099938\\
0.0154046250809462	0.96173411787193\\
0.0155314056706659	0.962198162200246\\
0.0156566331327613	0.962665986680842\\
0.0157802949445906	0.963137544531657\\
0.0159023787400757	0.963612788597301\\
0.0160228723109387	0.964091671353763\\
0.0161417636079229	0.964574144913165\\
0.0162590407419978	0.965060161028555\\
0.0163746919855476	0.965549671098726\\
0.0164887057735444	0.966042626173078\\
0.0166010707047045	0.966538976956515\\
0.0167117755426282	0.967038673814372\\
0.016820809216924	0.967541666777381\\
0.0169281608243155	0.968047905546663\\
0.0170338196297313	0.968557339498763\\
0.0171377750673789	0.969069917690712\\
0.0172400167418013	0.969585588865117\\
0.017340534428916	0.97010430145529\\
0.0174393180770383	0.970626003590404\\
0.0175363578078856	0.971150643100681\\
0.0176316439175657	0.971678167522607\\
0.017725166877547	0.972208524104179\\
0.0178169173356115	0.972741659810182\\
0.0179068861167898	0.973277521327488\\
0.0179950642242788	0.973816055070392\\
0.0180814428403412	0.97435720718597\\
0.0181660133271874	0.97490092355946\\
0.0182487672278392	0.975447149819679\\
0.0183296962669754	0.975995831344455\\
0.0184087923517596	0.976546913266093\\
0.0184860475726493	0.97710034047686\\
0.0185614542041867	0.977656057634497\\
0.0186350047057714	0.97821400916775\\
0.0187066917224148	0.978774139281931\\
0.0187765080854747	0.979336391964497\\
0.0188444468133732	0.979900710990646\\
0.0189105011122939	0.980467039928946\\
0.0189746643768621	0.981035322146977\\
0.0190369301908048	0.981605500816989\\
0.0190972923275925	0.98217751892159\\
0.0191557447510618	0.982751319259447\\
0.019212281616019	0.983326844451005\\
0.0192668972688249	0.983904036944223\\
0.0193195862479595	0.984482839020333\\
0.0193703432845689	0.985063192799611\\
0.0194191633029918	0.985645040247161\\
0.019466041421267	0.986228323178724\\
0.0195109729516217	0.986812983266493\\
0.0195539534009403	0.987398962044946\\
0.0195949784712137	0.987986200916694\\
0.0196340440599691	0.988574641158338\\
0.0196711462606801	0.989164223926346\\
0.0197062813631577	0.989754890262931\\
0.0197394458539208	0.990346581101953\\
0.0197706364165479	0.990939237274819\\
0.0197998499320089	0.991532799516408\\
0.0198270834789765	0.992127208470989\\
0.0198523343341187	0.992722404698162\\
0.0198755999723711	0.993318328678801\\
0.0198968780671892	0.993914920821004\\
0.0199161664907812	0.994512121466054\\
0.0199334633143209	0.995109870894384\\
0.0199487668081404	0.995708109331549\\
0.0199620754419029	0.996306776954205\\
0.0199733878847563	0.996905813896088\\
0.0199827030054656	0.997505160254003\\
0.0199900198725266	0.998104756093813\\
0.0199953377542586	0.998704541456434\\
0.0199986561188779	0.99930445636383\\
0.0199999746345508	0.999904440825011\\
0.0199992931694268	1.00050443484203\\
0.019996611791652	1.00110437841598\\
0.0199919307693617	1.00170421155302\\
0.0199852505706544	1.00230387427031\\
0.0199765718635444	1.0029033066021\\
0.0199658955158951	1.00350244860565\\
0.0199532225953324	1.00410124036726\\
0.0199385543691377	1.00469962200825\\
0.0199218923041216	1.00529753369096\\
0.0199032380664766	1.00589491562471\\
0.0198825935216109	1.00649170807181\\
0.0198599607339619	1.00708785135351\\
0.0198353419667893	1.00768328585598\\
0.0198087396819495	1.00827795203627\\
0.0197801565396487	1.00887179042826\\
0.0197495953981773	1.00946474164859\\
0.0197170593136241	1.01005674640265\\
0.0196825515395703	1.01064774549045\\
0.0196460755267646	1.01123767981257\\
0.019607634922778	1.01182649037608\\
0.0195672335716387	1.0124141183004\\
0.0195248755134482	1.01300050482324\\
0.019480564983977	1.01358559130643\\
0.0194343064142412	1.01416931924181\\
0.0193861044300592	1.01475163025708\\
0.0193359638515892	1.01533246612161\\
0.0192838896928473	1.01591176875231\\
0.019229887161206	1.01648948021939\\
0.0191739616568734	1.01706554275219\\
0.0191161187723533	1.01763989874493\\
0.0190563642918861	1.01821249076251\\
0.0189947041908699	1.01878326154618\\
0.0189311446352635	1.01935215401935\\
0.0188656919809695	1.01991911129324\\
0.0187983527731988	1.02048407667261\\
0.0187291337458159	1.02104699366138\\
0.0186580418206661	1.02160780596832\\
0.0185850841068825	1.02216645751267\\
0.0185102679001757	1.02272289242974\\
0.0184336006821041	1.0232770550765\\
0.0183550901193255	1.02382889003716\\
0.0182747440628309	1.02437834212866\\
0.0181925705471589	1.02492535640627\\
0.0181085777895926	1.02546987816902\\
0.0180227741893378	1.02601185296516\\
0.0179351683266829	1.02655122659769\\
0.0178457689621414	1.02708794512969\\
0.017754585035575	1.02762195488974\\
0.0176616256653005	1.02815320247732\\
0.0175669001471775	1.02868163476812\\
0.0174704179536788	1.02920719891934\\
0.0173721887329433	1.02972984237501\\
0.0172722223078113	1.03024951287122\\
0.017170528674842	1.03076615844134\\
0.017067118003314	1.03127972742126\\
0.0169620006342082	1.03179016845451\\
0.0168551870791739	1.0322974304974\\
0.0167466880194776	1.03280146282416\\
0.0166365143049349	1.03330221503197\\
0.0165246769528254	1.03379963704602\\
0.0164111871467912	1.03429367912454\\
0.0162960562357182	1.03478429186372\\
0.0161792957326017	1.03527142620271\\
0.0160609173133946	1.03575503342847\\
0.0159409328158402	1.03623506518068\\
0.0158193542382883	1.03671147345656\\
0.0156961937384954	1.0371842106157\\
0.0155714636324086	1.03765322938476\\
0.0154451763929349	1.03811848286226\\
0.0153173446486927	1.03857992452323\\
0.0151879811827502	1.03903750822391\\
0.0150570989313459	1.03949118820629\\
0.0149247109825961	1.03994091910277\\
0.0147908305751852	1.04038665594061\\
0.0146554710970424	1.04082835414652\\
0.0145186460840028	1.04126596955104\\
0.0143803692184536	1.04169945839299\\
0.0142406543279662	1.04212877732385\\
0.0140995153839132	1.04255388341208\\
0.0139569665000713	1.04297473414744\\
0.0138130219312102	1.04339128744519\\
0.0136676960716667	1.04380350165036\\
0.0135210034539058	1.04421133554186\\
0.013372958747067	1.04461474833666\\
0.0132235767554976	1.0450136996938\\
0.0130728724172722	1.04540814971848\\
0.0129208608026992	1.04579805896602\\
0.0127675571128132	1.04618338844584\\
0.0126129766778555	1.0465640996253\\
0.0124571349557408	1.0469401544336\\
0.0123000475305115	1.04731151526557\\
0.0121417301107791	1.04767814498544\\
0.0119821985281537	1.04804000693055\\
0.0118214687356606	1.04839706491499\\
0.0116595568061452	1.04874928323328\\
0.0114964789306654	1.04909662666386\\
0.0113322514168729	1.0494390604727\\
0.0111668906873822	1.04977655041668\\
0.0110004132781285	1.05010906274711\\
0.010832835836714	1.05043656421301\\
0.0106641751207431	1.05075902206453\\
0.0104944479961469	1.05107640405613\\
0.0103236714354965	1.0513886784499\\
0.0101518625163055	1.05169581401864\\
0.0099790384193228	1.05199778004907\\
0.00980521642681399	1.05229454634482\\
0.00963041392083347	1.05258608322951\\
0.00945464838148619	1.0528723615497\\
0.0092779373851796	1.0531533526778\\
0.00910029860286609	1.05342902851492\\
0.00892174979827585	1.05369936149371\\
0.00874230882614055	1.05396432458111\\
0.00856199363040787	1.05422389128102\\
0.00838082224244711	1.05447803563698\\
0.0081988127792461	1.05472673223478\\
0.0080159834415995	1.05496995620497\\
0.0078323525122887	1.05520768322534\\
0.0076479383542536	1.05543988952339\\
0.00746275940875635	1.05566655187869\\
0.00727683419353716	1.05588764762519\\
0.00709018130096262	1.05610315465351\\
0.00690281939616646	1.05631305141311\\
0.00671476721518301	1.05651731691449\\
0.00652604356307366	1.05671593073128\\
0.00633666731204636	1.05690887300227\\
0.00614665739956839	1.05709612443337\\
0.00595603282647266	1.05727766629961\\
0.00576481265505763	1.05745348044696\\
0.00557301600718108	1.05762354929413\\
0.00538066206234798	1.0577878558344\\
0.00518777005579252	1.05794638363724\\
0.00499435927655461	1.05809911685\\
0.00480044906555099	1.0582460401995\\
0.00460605881364117	1.05838713899351\\
0.00441120795968837	1.05852239912228\\
0.00421591598861559	1.05865180705991\\
0.0040202024294572	1.0587753498657\\
0.00382408685340602	1.05889301518549\\
0.00362758887185623	1.05900479125284\\
0.00343072813444223	1.05911066689023\\
0.00323352432707372	1.05921063151019\\
0.0030359971699671	1.05930467511635\\
0.00283816641567347	1.05939278830441\\
0.0026400518471034	1.05947496226314\\
0.00244167327554866	1.0595511887752\\
0.00224305053870108	1.05962146021801\\
0.00204420349866884	1.05968576956448\\
0.00184515203999023	1.05974411038373\\
0.00164591606764524	1.05979647684174\\
0.00144651550506508	1.05984286370189\\
0.00124697029213984	1.05988326632553\\
0.00104730038322452	1.05991768067245\\
0.000847525745143622	1.05994610330123\\
0.000647666355194483	1.05996853136964\\
0.000447742199149545	1.05998496263488\\
0.000247773269257809	1.05999539545385\\
4.77795622456239e-05	1.05999982878326\\
-0.000152218922682969	1.0599982621798\\
-0.000352202185846143	1.05999069580012\\
-0.000552150229084233	1.05997713040085\\
-0.000752043057759538	1.05995756733852\\
-0.000951860682755768	1.05993200856942\\
-0.00115158312247694	1.05990045664941\\
-0.00135119040484551	1.05986291473365\\
-0.00155066256929958	1.05981938657631\\
-0.00174997966878894	1.05976987653015\\
-0.00194912177176973	1.05971438954615\\
-0.00214806896419761	1.05965293117296\\
-0.00234680135151912	1.05958550755636\\
-0.00254529906066113	1.05951212543865\\
-0.00274354224201816	1.059432792158\\
-0.00294151107143726	1.05934751564766\\
-0.00313918575220048	1.05925630443521\\
-0.00333654651700444	1.0591591676417\\
-0.00353357362993716	1.05905611498072\\
-0.00373024738845151	1.05894715675746\\
-0.00392654812533555	1.05883230386764\\
-0.00412245621067917	1.05871156779647\\
-0.00431795205383709	1.05858496061744\\
-0.00451301610538791	1.05845249499117\\
-0.00470762885908903	1.05831418416411\\
-0.00490177085382724	1.05817004196723\\
-0.00509542267556487	1.05802008281463\\
-0.00528856495928111	1.05786432170209\\
-0.00548117839090855	1.05770277420561\\
-0.00567324370926453	1.05753545647979\\
-0.00586474170797727	1.05736238525627\\
-0.00605565323740649	1.05718357784202\\
-0.00624595920655832	1.05699905211765\\
-0.00643564058499444	1.05680882653556\\
-0.00662467840473509	1.05661292011815\\
-0.0068130537621558	1.05641135245592\\
-0.00700074781987783	1.05620414370544\\
-0.0071877418086518	1.05599131458744\\
-0.00737401702923467	1.05577288638463\\
-0.00755955485425962	1.05554888093966\\
-0.00774433673009875	1.05531932065289\\
-0.00792834417871846	1.05508422848016\\
-0.00811155879952722	1.05484362793047\\
-0.00829396227121565	1.0545975430637\\
-0.00847553635358857	1.05434599848812\\
-0.00865626288938904	1.05408901935798\\
-0.00883612380611411	1.05382663137098\\
-0.00901510111782199	1.05355886076569\\
-0.00919317692693064	1.05328573431897\\
-0.00937033342600754	1.05300727934321\\
-0.00954655289955044	1.05272352368369\\
-0.00972181772575881	1.05243449571574\\
-0.0098961103782961	1.05214022434191\\
-0.0100694134280423	1.0518407389891\\
-0.0102417095448368	1.0515360696056\\
-0.0104129814992116	1.05122624665807\\
-0.0105832121641139	1.05091130112857\\
-0.0107523845166191	1.05059126451139\\
-0.010920481639633	1.05026616880991\\
-0.0110874867235832	1.04993604653343\\
-0.0112533830681007	1.04960093069392\\
-0.0114181540836891	1.04926085480267\\
-0.0115817832933842	1.048915852867\\
-0.0117442543344015	1.0485659593868\\
-0.0119055509597721	1.04821120935113\\
-0.0120656570399681	1.0478516382347\\
-0.0122245565645147	1.04748728199433\\
-0.012382233643592	1.04711817706533\\
-0.0125386725096234	1.04674436035788\\
-0.0126938575188527	1.04636586925336\\
-0.0128477731529083	1.04598274160055\\
-0.013000404020355	1.04559501571189\\
-0.0131517348582334	1.04520273035966\\
-0.0133017505335857	1.04480592477207\\
-0.0134504360449693	1.04440463862933\\
-0.0135977765239571	1.04399891205974\\
-0.013743757236624	1.0435887856356\\
-0.0138883635850203	1.04317430036923\\
-0.0140315811086317	1.0427554977088\\
-0.0141733954858252	1.04233241953422\\
-0.0143137925352812	1.04190510815297\\
-0.0144527582174118	1.04147360629583\\
-0.0145902786357646	1.04103795711261\\
-0.0147263400384124	1.04059820416789\\
-0.0148609288193282	1.04015439143658\\
-0.0149940315197462	1.03970656329958\\
-0.0151256348295071	1.03925476453935\\
-0.0152557255883898	1.03879904033537\\
-0.0153842907874267	1.03833943625969\\
-0.015511317570205	1.03787599827234\\
-0.0156367932341524	1.03740877271672\\
-0.015760705231807	1.03693780631501\\
-0.0158830411720722	1.03646314616344\\
-0.0160037888214561	1.03598483972764\\
-0.0161229361052943	1.03550293483786\\
-0.0162404711089577	1.03501747968418\\
-0.0163563820790439	1.03452852281171\\
-0.0164706574245524	1.03403611311574\\
-0.016583285718044	1.03354029983681\\
-0.0166942556967832	1.03304113255586\\
-0.0168035562638645	1.03253866118918\\
-0.0169111764893223	1.0320329359835\\
-0.0170171056112239	1.03152400751091\\
-0.0171213330367451	1.03101192666385\\
-0.0172238483432304	1.03049674464996\\
-0.0173246412792346	1.02997851298701\\
-0.017423701765548	1.02945728349776\\
-0.0175210198962045	1.0289331083047\\
-0.0176165859394722	1.02840603982492\\
-0.0177103903388264	1.02787613076483\\
-0.0178024237139053	1.02734343411489\\
-0.0178926768614482	1.02680800314433\\
-0.0179811407562154	1.02626989139579\\
-0.0180678065518912	1.02572915268\\
-0.0181526655819683	1.02518584107039\\
-0.0182357093606143	1.02464001089766\\
-0.0183169295835207	1.02409171674437\\
-0.0183963181287328	1.02354101343949\\
-0.0184738670574622	1.02298795605287\\
-0.0185495686148807	1.02243259988981\\
-0.0186234152308957	1.02187500048546\\
-0.018695399520907	1.02131521359928\\
-0.0187655142865457	1.02075329520951\\
-0.0188337525163936	1.02018930150751\\
-0.0189001073866846	1.01962328889218\\
-0.0189645722619869	1.01905531396432\\
-0.0190271406958668	1.01848543352094\\
-0.0190878064315331	1.01791370454962\\
-0.0191465634024626	1.01734018422276\\
-0.0192034057330073	1.01676492989194\\
-0.0192583277389815	1.01618799908209\\
-0.0193113239282304	1.01560944948583\\
-0.0193623890011791	1.01502933895762\\
-0.019411517851363	1.01444772550805\\
-0.019458705565938	1.01386466729796\\
-0.0195039474261719	1.0132802226327\\
-0.0195472389079164	1.01269444995624\\
-0.0195885756820594	1.01210740784537\\
-0.019627953614958	1.0115191550038\\
-0.0196653687688517	1.01092975025633\\
-0.0197008174022562	1.01033925254294\\
-0.0197342959703378	1.00974772091291\\
-0.0197658011252676	1.00915521451891\\
-0.0197953297165563	1.0085617926111\\
-0.0198228787913694	1.00796751453115\\
-0.0198484455948223	1.00737243970639\\
-0.0198720275702561	1.00677662764381\\
-0.0198936223594929	1.00618013792411\\
-0.0199132278030716	1.00558303019576\\
-0.0199308419404643	1.00498536416905\\
-0.019946463010272	1.00438719961008\\
-0.0199600894504007	1.0037885963348\\
-0.0199717198982176	1.00318961420304\\
-0.0199813531906878	1.00259031311252\\
-0.01998898836449	1.00199075299285\\
-0.0199946246561132	1.00139099379954\\
-0.0199982615019328	1.00079109550801\\
-0.0199998985382675	1.00019111810759\\
};
\addplot[only marks,mark=x,mark options={},mark size=1.5000pt,draw=black,fill=black] plot table[row sep=crcr]{%
0	1\\
};
\addplot[only marks,mark=*,mark options={},mark size=1.5000pt,draw=black,fill=black] plot table[row sep=crcr]{%
0	0\\
};
\end{axis}
\end{tikzpicture}%

%% file: sensorConeSonar.tex
%
%
\begin{tikzpicture}

\begin{axis}[%
width=0.75\figurewidth,
height=\figureheight,
at={(0\figurewidth,0\figureheight)},
scale only axis,
xmin=-0.7,
xmax=0.7,
xtick={-0.6,-0.4,-0.2,0,0.2,0.4,0.6},
xticklabels={\empty},
xmajorgrids,
ymin=0,
ymax=1.4,
ytick={0,0.2,0.4,0.6,0.8,1,1.2,1.4},
yticklabels={\empty},
ymajorgrids,
axis background/.style={fill=white}
]
\addplot [color=black,solid,line width=1.0pt,forget plot]
  table[row sep=crcr]{%
-0.5	0.894427190999916\\
-0.499975000208333	0.893899480581951\\
-0.499900003333289	0.893389702920077\\
-0.499775016874494	0.892897916502991\\
-0.499600053330489	0.892424175876055\\
-0.499375130197483	0.891968531636395\\
-0.499100269967602	0.891531030426025\\
-0.49877550012664	0.891111714923008\\
-0.49840085315131	0.890710623830605\\
-0.497976366505997	0.890327791864425\\
-0.497502082639013	0.889963249737546\\
-0.496978048978349	0.889617024143612\\
-0.496404317926933	0.889289137737883\\
-0.495780946857394	0.888979609116253\\
-0.495107998106319	0.888688452792221\\
-0.494385538968021	0.888415679171811\\
-0.493613641687814	0.888161294526464\\
-0.49279238345478	0.887925300963883\\
-0.491921846394061	0.88770769639686\\
-0.491002117558635	0.887508474510082\\
-0.490033288920621	0.887327624724938\\
-0.489015457362074	0.88716513216234\\
-0.487948724665303	0.88702097760358\\
-0.486833197502687	0.886895137449247\\
-0.485668987426015	0.886787583676228\\
-0.484456210855322	0.88669828379282\\
-0.483194989067257	0.886627200792001\\
-0.481885448182945	0.886574293102875\\
-0.480527719155386	0.886539514540349\\
-0.479121937756348	0.886522814253085\\
-0.477668244562803	0.886524136669761\\
-0.476166784942857	0.886543421443718\\
-0.474617709041221	0.886580603396027\\
-0.473021171764194	0.886635612457054\\
-0.471377332764173	0.88670837360658\\
-0.469686356423689	0.886798806812552\\
-0.467948411838967	0.886906826968539\\
-0.466163672803017	0.887032343829975\\
-0.464332317788255	0.887175261949275\\
-0.462454529928656	0.887335480609911\\
-0.460530497001442	0.887512893759561\\
-0.458560411408303	0.887707389942408\\
-0.456544470156154	0.887918852230724\\
-0.454482874837443	0.888147158155834\\
-0.452375831609982	0.888392179638596\\
-0.450223551176338	0.88865378291951\\
-0.448026248762762	0.888931828488603\\
-0.445784144097664	0.889226171015215\\
-0.443497461389642	0.889536659277851\\
-0.441166429305061	0.889863136094234\\
-0.438791280945186	0.890205438251732\\
-0.436372253822876	0.890563396438324\\
-0.433909589838825	0.890936835174273\\
-0.431403535257381	0.891325572744692\\
-0.428854340681912	0.891729421133197\\
-0.426262261029753	0.892148185956836\\
-0.423627555506708	0.892581666402491\\
-0.420950487581134	0.893029655164985\\
-0.418231324957593	0.893491938387085\\
-0.415470339550082	0.893968295601648\\
-0.412667807454839	0.89445849967613\\
-0.40982400892274	0.894962316759696\\
-0.406939228331267	0.895479506233189\\
-0.404013754156076	0.8960098206622\\
-0.401047878942146	0.896553005753498\\
-0.398041899274528	0.897108800315093\\
-0.394996115748683	0.897676936220208\\
-0.391910832940425	0.898257138375416\\
-0.388786359375464	0.898849124693256\\
-0.385623007498553	0.899452606069597\\
-0.382421093642244	0.900067286366044\\
-0.379180937995254	0.900692862397702\\
-0.375902864570448	0.90132902392658\\
-0.372587201172435	0.90197545366095\\
-0.369234279364794	0.902631827260977\\
-0.365844434436911	0.903297813350918\\
-0.362418005370453	0.903973073538212\\
-0.358955334805472	0.904657262439771\\
-0.355456769006139	0.905350027715796\\
-0.351922657826118	0.906051010111408\\
-0.348353354673583	0.906759843506442\\
-0.344749216475874	0.90747615497368\\
-0.341110603643807	0.908199564845847\\
-0.337437880035634	0.908929686791678\\
-0.333731412920654	0.90966612790133\\
-0.329991572942491	0.910408488781462\\
-0.326218734082026	0.911156363660235\\
-0.322413273620001	0.911909340502528\\
-0.31857557209929	0.912667001135627\\
-0.314706013286848	0.913428921385633\\
-0.310804984135332	0.914194671224848\\
-0.306872874744406	0.91496381493035\\
-0.302910078321732	0.915735911253967\\
-0.298916991143649	0.916510513603865\\
-0.294894012515549	0.917287170237905\\
-0.290841544731942	0.918065424468935\\
-0.286759993036228	0.918844814882155\\
-0.282649765580177	0.919624875564673\\
-0.278511273383109	0.920405136347329\\
-0.274344930290794	0.921185123058866\\
-0.27015115293407	0.921964357792481\\
-0.265930360687178	0.922742359184766\\
-0.261682975625825	0.923518642707022\\
-0.257409422484978	0.924292720968897\\
-0.253110128616389	0.925064104034261\\
-0.248785523945864	0.92583229974922\\
-0.244436040930264	0.926596814082097\\
-0.240062114514267	0.927357151475215\\
-0.23566418208687	0.928112815208254\\
-0.231242683437651	0.928863307772925\\
-0.226798060712789	0.929608131258673\\
-0.222330758370853	0.930346787749062\\
-0.217841223138356	0.93107877972848\\
-0.213329903965079	0.931803610498751\\
-0.208797251979179	0.932520784605201\\
-0.204243720442079	0.933229808271685\\
-0.199669764703137	0.933930189844042\\
-0.195075842154115	0.934621440241417\\
-0.190462412183441	0.935303073414828\\
-0.185829936130267	0.935974606812343\\
-0.181178877238337	0.936635561850167\\
-0.176509700609665	0.937285464388929\\
-0.171822873158024	0.937923845214403\\
-0.167118863562251	0.938550240521878\\
-0.162398142219388	0.939164192403339\\
-0.157661181197634	0.939765249336627\\
-0.152908454189145	0.940352966675658\\
-0.148140436462659	0.940926907140828\\
-0.143357604815978	0.941486641308638\\
-0.138560437528279	0.942031748099594\\
-0.133749414312294	0.942561815263399\\
-0.128925016266335	0.943076439860453\\
-0.124087725826186	0.943575228738634\\
-0.119238026716862	0.944057799004361\\
-0.11437640390423	0.944523778486897\\
-0.109503343546521	0.944972806194877\\
-0.10461933294571	0.94540453276402\\
-0.0997248604987864	0.945818620895\\
-0.0948204156489172	0.946214745780464\\
-0.0899064888364998	0.946592595520184\\
-0.0849835714501205	0.946951871523339\\
-0.0800521557774158	0.947292288896965\\
-0.0751127349558429	0.947613576819617\\
-0.0701658029233685	0.947915478899302\\
-0.0652118543690729	0.948197753514798\\
-0.0602513846836833	0.948460174139486\\
-0.0552848899100348	0.948702529646869\\
-0.0503128666934658	0.948924624596985\\
-0.0453358122321549	0.949126279502968\\
-0.0403542242274003	0.949307331077069\\
-0.0353686008338515	0.949467632455469\\
-0.030379440609693	0.949607053401301\\
-0.0253872424667895	0.949725480485329\\
-0.0203925056207956	0.949822817243814\\
-0.0153957295412331	0.949898984313119\\
-0.0103974139015463	0.949953919540708\\
-0.00539805852913365	0.949987578072228\\
-0.000398163355366618	0.949999932414443\\
0.0046017716344042	0.949990972473843\\
0.0096012464508464	0.949960705570822\\
0.0145997611506445	0.949909156429397\\
0.019596815886494	0.949836367142483\\
0.0245919109570853	0.94974239711282\\
0.0295845468570743	0.949627322969731\\
0.0345742243270311	0.949491238461906\\
0.0395604444033671	0.949334254326537\\
0.0445427084682297	0.949156498135128\\
0.0495205182993641	0.948958114116423\\
0.0544933761199355	0.948739262956901\\
0.0594607846483062	0.9485001215794\\
0.0644222471477624	0.948240882900429\\
0.0693772674761889	0.947961755566834\\
0.0743253501356819	0.947662963672498\\
0.0792660003220989	0.947344746455811\\
0.0841987239745385	0.947007357978704\\
0.0891230278247462	0.946651066788054\\
0.0940384194464402	0.94627615556034\\
0.0989444073045546	0.945882920730423\\
0.103840500804392	0.945471672105393\\
0.108726210340682	0.945042732464416\\
0.113601047346544	0.944596437145573\\
0.118464524342337	0.944133133620663\\
0.123316154984417	0.943653181058992\\
0.128155454113761	0.943156949881158\\
0.13298193780449	0.942644821303862\\
0.137795123412256	0.942117186876768\\
0.142594529622511	0.941574448012458\\
0.147379676498631	0.941017015510482\\
0.152150085529917	0.940445309076547\\
0.156905279679441	0.93985975683785\\
0.161644783431752	0.939260794855538\\
0.166368122840423	0.93864886663529\\
0.171074825575449	0.938024422636984\\
0.17576442097048	0.937387919784374\\
0.180436440069884	0.936739820975711\\
0.185090415675643	0.936080594596183\\
0.189725882394077	0.935410714033046\\
0.194342376682376	0.934730657194262\\
0.198939436894958	0.93404090603146\\
0.203516603329633	0.933341946067967\\
0.208073418273571	0.932634265932645\\
0.212609426049076	0.931918356900229\\
0.21712417305915	0.931194712438804\\
0.221617207832855	0.930463827765064\\
0.226088081070456	0.929726199407899\\
0.230536345688356	0.928982324780875\\
0.234961556863801	0.92823270176408\\
0.23936327207936	0.927477828295817\\
0.24374105116718	0.926718201974544\\
0.248094456352999	0.925954319671451\\
0.252423052299929	0.925186677154006\\
0.25672640615198	0.92441576872078\\
0.261004087577354	0.923642086847802\\
0.265255668811473	0.922866121846686\\
0.269480724699756	0.922088361534703\\
0.273678832740136	0.92130929091696\\
0.277849573125306	0.920529391880809\\
0.281992528784705	0.919749142902566\\
0.286107285426219	0.918969018766597\\
0.290193431577611	0.918189490296795\\
0.294250558627673	0.917411024100437\\
0.29827826086708	0.916634082324397\\
0.302276135528965	0.915859122423645\\
0.306243782829193	0.915086596941944\\
0.31018080600634	0.914316953304647\\
0.31408681136137	0.913550633623444\\
0.317961408297001	0.912788074512918\\
0.321804209356771	0.912029706918726\\
0.325614830263773	0.911275955957229\\
0.329392889959094	0.910527240766337\\
0.333138010639912	0.909783974367365\\
0.336849817797281	0.909046563537658\\
0.340527940253576	0.908315408693722\\
0.344172010199619	0.907590903784613\\
0.347781663231451	0.906873436195304\\
0.351356538386777	0.906163386659741\\
0.35489627818106	0.905461129183317\\
0.358400528643272	0.904767030974449\\
0.361868939351284	0.904081452384962\\
0.365301163466919	0.903404746858986\\
0.368696857770623	0.902737260890033\\
0.372055682695796	0.902079333985964\\
0.375377302362745	0.901431298641518\\
0.378661384612272	0.9007934803181\\
0.381907601038887	0.900166197430498\\
0.385115627023654	0.899549761340235\\
0.388285141766647	0.89894447635523\\
0.391415828319033	0.898350639735463\\
0.394507373614766	0.897768541704338\\
0.397559468501892	0.897198465465439\\
0.400571807773467	0.896640687224386\\
0.403544090198073	0.896095476215483\\
0.406476018549945	0.895563094732882\\
0.409367299638691	0.895043798165971\\
0.412217644338611	0.894537835038706\\
0.415026767617611	0.894045447052622\\
0.417794388565704	0.893566869133237\\
0.420520230423101	0.893102329479618\\
0.423204020607888	0.892652049616825\\
0.425845490743283	0.892216244451005\\
0.428444376684474	0.891795122326887\\
0.431000418545032	0.89138888508745\\
0.433513360722901	0.890997728135535\\
0.435982951925958	0.890621840497178\\
0.438408945197141	0.890261404886464\\
0.440791097939143	0.889916597771685\\
0.443129171938676	0.889587589442612\\
0.445422933390288	0.889274544078694\\
0.447672152919746	0.888977619817986\\
0.449876605606971	0.888696968826656\\
0.452036071008531	0.888432737368876\\
0.454150333179685	0.888185065876952\\
0.456219180695979	0.887954089021537\\
0.458242406674385	0.887739935781771\\
0.46021980879399	0.887542729515216\\
0.462151189316232	0.887362588027444\\
0.464036355104666	0.887199623641161\\
0.465875117644286	0.887053943264727\\
0.46766729306037	0.886925648459976\\
0.469412702136868	0.886814835509217\\
0.471111170334329	0.886721595481303\\
0.472762527807348	0.886646014296693\\
0.474366609421554	0.886588172791394\\
0.475923254770121	0.886548146779701\\
0.477432308189813	0.886526007115659\\
0.478893618776545	0.886521819753167\\
0.480307040400476	0.886535645804646\\
0.481672431720622	0.886567541598216\\
0.482989656198987	0.886617558733304\\
0.484258582114223	0.886685744134647\\
0.485479082574795	0.886772140104607\\
0.486651035531674	0.886876784373777\\
0.487774323790541	0.886999710149816\\
0.488848835023507	0.887140946164478\\
0.489874461780342	0.887300516718787\\
0.490851101499227	0.887478441726344\\
0.491778656517003	0.887674736754718\\
0.492657034078942	0.887889413064897\\
0.493486146348019	0.888122477648781\\
0.494265910413698	0.888373933264695\\
0.494996248300223	0.8886437784709\\
0.495677086974413	0.888932007657092\\
0.496308358352969	0.889238611073873\\
0.496889999309278	0.889563574860191\\
0.49742195167973	0.889906881068742\\
0.497904162269531	0.890268507689323\\
0.498336582858023	0.890648428670146\\
0.498719170203509	0.891046613937103\\
0.499051886047573	0.891463029411002\\
0.499334697118907	0.891897637022764\\
0.49956757513664	0.892350394726596\\
0.499750496813164	0.892821256511161\\
0.499883443856464	0.893310172408734\\
0.499966402971947	0.893817088502392\\
0.49999936586377	0.894341946931224\\
0.499982329235671	0.894884685893607\\
0.499915294791299	0.895445239648547\\
0.499798269234043	0.896023538515127\\
0.49963126426636	0.896619508870082\\
0.499414296588609	0.897233073143523\\
0.499147387897376	0.897864149812848\\
0.498830564883309	0.898512653394874\\
0.498463859228443	0.899178494436209\\
0.49804730760304	0.899861579501924\\
0.497580951661915	0.900561811162543\\
0.497064838040273	0.901279087979396\\
0.496499018349046	0.902013304488393\\
0.495883549169733	0.902764351182236\\
0.495218492048737	0.903532114491145\\
0.494503913491216	0.904316476762122\\
0.493739884954432	0.90511731623682\\
0.492926482840601	0.905934507028069\\
0.492063788489257	0.906767919095104\\
0.491151888169116	0.907617418217563\\
0.490190873069449	0.908482865968317\\
0.489180839290967	0.909364119685178\\
0.488121887836205	0.910261032441579\\
0.487014124599426	0.911173453016254\\
0.485857660356031	0.912101225862034\\
0.48465261075148	0.913044191073785\\
0.48339909628973	0.914002184355603\\
0.482097242321183	0.914975036987311\\
0.480747179030149	0.915962575790366\\
0.479349041421834	0.916964623093233\\
0.477902969308833	0.917980996696333\\
0.476409107297152	0.91901150983664\\
0.474867604771748	0.920055971152025\\
0.473278615881588	0.921114184645431\\
0.471642299524238	0.922185949648993\\
0.469958819329969	0.923271060788184\\
0.468228343645398	0.9243693079461\\
0.466451045516652	0.925480476227987\\
0.464627102672062	0.926604345926121\\
0.462756697504392	0.927740692485141\\
0.460840017052602	0.928889286467968\\
0.458877252983138	0.930049893522409\\
0.456868601570772	0.931222274348575\\
0.454814263678972	0.932406184667236\\
0.452714444739815	0.933601375189236\\
0.450569354733444	0.9348075915861\\
0.448379208167073	0.936024574461959\\
0.446144224053534	0.937252059326933\\
0.443864625889375	0.938489776572106\\
0.441540641632513	0.939737451446238\\
0.439172503679437	0.940994804034347\\
0.436760448841969	0.942261549238312\\
0.434304718323582	0.943537396759646\\
0.431805557695283	0.944822051084587\\
0.429263216871051	0.946115211471653\\
0.42667795008285	0.947416571941827\\
0.424050015855204	0.948725821271519\\
0.421379676979347	0.950042642988461\\
0.41866720048694	0.951366715370704\\
0.415912857623373	0.952697711448875\\
0.413116923820636	0.954035299011842\\
0.41027967866978	0.955379140615973\\
0.407401405892956	0.956728893598137\\
0.404482393315043	0.958084210092615\\
0.401522932834865	0.959444737052093\\
0.398523320396006	0.960810116272882\\
0.395483855957208	0.96217998442456\\
0.392404843462384	0.963553973084169\\
0.389286590810216	0.964931708775151\\
0.386129409823372	0.966312813011178\\
0.382933616217319	0.967696902345025\\
0.379699529568754	0.969083588422675\\
0.376427473283648	0.970472478042765\\
0.373117774564901	0.971863173221576\\
0.369770764379629	0.973255271263678\\
0.36638677742606	0.974648364838395\\
0.36296615210007	0.97604204206223\\
0.35950923046134	0.977435886587378\\
0.356016358199155	0.978829477696475\\
0.352487884597829	0.980222390403691\\
0.348924162501782	0.981614195562305\\
0.345325548280254	0.983004459978864\\
0.341692401791668	0.984392746534041\\
0.338025086347646	0.985778614310277\\
0.334323968676676	0.987161618726324\\
0.33058941888744	0.988541311678743\\
0.326821810431806	0.989917241690452\\
0.323021520067479	0.991288954066368\\
0.319188927820329	0.992655991056221\\
0.315324416946388	0.994017892024553\\
0.311428373893521	0.995374193627956\\
0.307501188262787	0.996724429999557\\
0.303543252769477	0.998068132940761\\
0.299554963203843	0.999404832120234\\
0.295536718391516	1.00073405528013\\
0.291488920153629	1.00205532844948\\
0.287411973266634	1.0033681761648\\
0.283306285421822	1.00467212169765\\
0.279172267184555	1.00596668728936\\
0.275010331953212	1.00725139439253\\
0.270820895917849	1.00852576391939\\
0.266604378018577	1.00978931649682\\
0.262361199903673	1.0110415727279\\
0.258091785887412	1.01228205345983\\
0.253796562907638	1.01351028005802\\
0.24947596048307	1.01472577468618\\
0.24513041067035	1.01592806059219\\
0.240760348020837	1.01711666239956\\
0.236366209537155	1.01829110640412\\
0.23194843462949	1.01945092087578\\
0.227507465071652	1.02059563636507\\
0.223043744956896	1.02172478601408\\
0.218557720653514	1.02283790587159\\
0.214049840760197	1.02393453521193\\
0.209520556061178	1.02501421685736\\
0.204970319481153	1.02607649750348\\
0.200399586039988	1.02712092804735\\
0.195808812807217	1.02814706391794\\
0.19119845885634	1.02915446540839\\
0.186568985218909	1.0301426980098\\
0.181920854838429	1.031111332746\\
0.177254532524066	1.03205994650886\\
0.172570484904161	1.03298812239374\\
0.167869180379575	1.03389545003452\\
0.163151089076842	1.03478152593779\\
0.158416682801159	1.0356459538156\\
0.15366643498921	1.03648834491648\\
0.148900820661816	1.03730831835391\\
0.144120316376441	1.03810550143202\\
0.139325400179527	1.03887952996779\\
0.1345165515587	1.0396300486094\\
0.129694251394813	1.04035671114994\\
0.124858981913865	1.04105918083634\\
0.120011226638775	1.04173713067259\\
0.115151470341029	1.04239024371704\\
0.110280198992209	1.04301821337301\\
0.10539789971539	1.04362074367247\\
0.10050506073643	1.04419754955197\\
0.0956021713351504	1.04474835712061\\
0.0906897217964057	1.04527290391931\\
0.0857682033610557	1.04577093917111\\
0.0808381081768429	1.04624222402189\\
0.0758999292491774	1.04668653177108\\
0.0709541603918366	1.04710364809202\\
0.066001296177585	1.04749337124138\\
0.0610418318887163	1.04785551225739\\
0.0560762634675271	1.0481898951464\\
0.051105087466721	1.0484963570574\\
0.0461288009997557	1.04877474844429\\
0.041147901691131	1.04902493321535\\
0.036162887626627	1.04924678886981\\
0.031174257303496	1.04944020662118\\
0.0261825095806128	1.04960509150706\\
0.0211881436285907	1.04974136248521\\
0.016191658879862	1.04984895251576\\
0.0111935549787387	1.04992780862933\\
0.00619433173144518	1.04997789198091\\
0.00119448905614057	1.04999917788943\\
-0.00380547306707426	1.04999165586289\\
-0.00880505464615373	1.04995532960913\\
-0.013803755727106	1.04989021703198\\
-0.0188010764439885	1.04979635021301\\
-0.0237965170688942	1.0496737753789\\
-0.0287895780619236	1.04952255285432\\
-0.0337797601211377	1.0493427570006\\
-0.0387665642324896	1.04913447614031\\
-0.0437494917197235	1.04889781246768\\
-0.0487280442942433	1.04863288194544\\
-0.0537017241049402	1.0483398141878\\
-0.0586700337879781	1.0480187523303\\
-0.0636324765165286	1.04766985288637\\
-0.0685885560504541	1.04729328559119\\
-0.0735377767859317	1.04688923323295\\
-0.0784796438050119	1.04645789147202\\
-0.083413662925111	1.04599946864825\\
-0.088339340748429	1.04551418557683\\
-0.0932561847112879	1.04500227533311\\
-0.098163703133389	1.0444639830268\\
-0.103061405266979	1.04389956556595\\
-0.107948801345927	1.04330929141119\\
-0.112825402634698	1.04269344032067\\
-0.117690721477226	1.04205230308622\\
-0.122544271345681	1.04138618126109\\
-0.127385566889122	1.04069538688\\
-0.132214123982028	1.03998024217171\\
-0.137029459772714	1.0392410792649\\
-0.141831092731613	1.03847823988765\\
-0.146618542699432	1.03769207506124\\
-0.151391330935162	1.03688294478862\\
-0.156148980163958	1.03605121773824\\
-0.160891014624861	1.03519727092356\\
-0.165616960118377	1.03432148937906\\
-0.170326344053895	1.03342426583291\\
-0.175018695496946	1.03250600037722\\
-0.179693545216295	1.03156710013595\\
-0.184350425730867	1.03060797893144\\
-0.188988871356491	1.02962905694963\\
-0.193608418252469	1.02863076040483\\
-0.198208604467962	1.02761352120422\\
-0.20278896998818	1.02657777661281\\
-0.207349056780391	1.02552396891911\\
-0.211888408839714	1.02445254510207\\
-0.216406572234726	1.02336395649974\\
-0.220903095152853	1.02225865847996\\
-0.22537752794555	1.02113711011363\\
-0.229829423173266	1.01999977385081\\
-0.234258335650189	1.01884711520017\\
-0.238663822488761	1.01767960241199\\
-0.243045443143971	1.01649770616521\\
-0.247402759457403	1.01530189925874\\
-0.251735335701057	1.01409265630737\\
-0.256042738620921	1.01287045344265\\
-0.26032453748029	1.01163576801885\\
-0.264580304102848	1.01038907832444\\
-0.268809612915478	1.0091308632992\\
-0.273012040990824	1.00786160225722\\
-0.277187168089581	1.00658177461601\\
-0.281334576702517	1.0052918596319\\
-0.285453852092227	1.00399233614186\\
-0.289544582334605	1.00268368231199\\
-0.293606358360037	1.00136637539265\\
-0.297638773994303	1.00004089148055\\
-0.301641425999202	0.998707705287737\\
-0.305613914112868	0.997367289917638\\
-0.3095558410898	0.996020116648238\\
-0.313466812740585	0.994666654722415\\
-0.317346437971318	0.993307371145498\\
-0.321194328822707	0.991942730490055\\
-0.325010100508876	0.990573194707948\\
-0.328793371455835	0.989199222949633\\
-0.332543763339642	0.9878212713907\\
-0.336260901124233	0.986439793065636\\
-0.339944413098929	0.985055237708761\\
-0.343593930915601	0.983668051602295\\
-0.347209089625508	0.982278677431503\\
-0.350789527715793	0.980887554146839\\
-0.35433488714563	0.979495116833015\\
-0.357844813382031	0.978101796584917\\
-0.361318955435296	0.976708020390253\\
-0.364756965894116	0.975314211018851\\
-0.36815850096031	0.973920786918481\\
-0.371523220483205	0.972528162117093\\
-0.374850787993654	0.971136746131335\\
-0.378140870737678	0.969746943881243\\
-0.381393139709744	0.96835915561094\\
-0.384607269685666	0.966973776815231\\
-0.387782939255125	0.96559119817193\\
-0.39091983085381	0.964211805479783\\
-0.394017630795174	0.962835979601828\\
-0.397076029301806	0.961464096414047\\
-0.400094720536403	0.960096526759143\\
-0.403073402632358	0.958733636405285\\
-0.406011777723943	0.957375786009668\\
-0.408909551976098	0.956023331086717\\
-0.411766435613811	0.954676621980771\\
-0.414582142951101	0.953336003843095\\
-0.41735639241958	0.952001816613035\\
-0.420088906596613	0.95067439500317\\
-0.422779412233059	0.949354068488291\\
-0.425427640280596	0.948041161298034\\
-0.428033325918628	0.946735992413012\\
-0.430596208580761	0.94543887556429\\
-0.433116031980864	0.944150119236019\\
-0.435592544138699	0.942870026671097\\
-0.438025497405112	0.941598895879676\\
-0.440414648486805	0.940337019650369\\
-0.44275975847066	0.939084685564001\\
-0.445060592847633	0.937842176009747\\
-0.447316921536204	0.936609768203513\\
-0.449528518905384	0.935387734208402\\
-0.45169516379728	0.934176340957132\\
-0.453816639549207	0.932975850276243\\
-0.455892734015358	0.931786518911975\\
-0.457923239588018	0.930608598557664\\
-0.45990795321832	0.929442335882516\\
-0.461846676436555	0.928287972561645\\
-0.463739215372018	0.927145745307221\\
-0.465585380772392	0.92601588590063\\
-0.467384988022675	0.924898621225483\\
-0.469137857163641	0.923794173301396\\
-0.470843812909839	0.922702759318385\\
-0.472502684667114	0.921624591671785\\
-0.474114306549673	0.92055987799757\\
-0.475678517396671	0.919508821207962\\
-0.477195160788327	0.918471619527235\\
-0.478664085061566	0.917448466527593\\
-0.480085143325183	0.916439551165036\\
-0.481458193474538	0.915445057815105\\
-0.482783098205759	0.914465166308419\\
-0.484059725029477	0.913500051965909\\
-0.485287946284075	0.912549885633652\\
-0.486467639148449	0.911614833717236\\
-0.487598685654296	0.910695058215549\\
-0.48868097269791	0.909790716753932\\
-0.489714392051486	0.908901962616604\\
-0.49069884037395	0.908028944778282\\
-0.491634219221292	0.907171807934939\\
-0.492520435056406	0.906330692533602\\
-0.493357399258446	0.905505734801152\\
-0.49414502813169	0.90469706677203\\
-0.494883242913907	0.903904816314811\\
-0.495571969784234	0.903129107157568\\
-0.496211139870558	0.902370058911974\\
-0.496800689256403	0.901627787096079\\
-0.497340558987322	0.900902403155726\\
-0.49783069507679	0.900194014484522\\
-0.498271048511609	0.899502724442344\\
-0.498661575256801	0.89882863237232\\
-0.499002236260016	0.898171833616236\\
-0.499292997455441	0.897532419528325\\
-0.499533829767196	0.896910477487417\\
-0.49972470911225	0.896306090907372\\
-0.499865616402829	0.895719339245801\\
-0.499956537548321	0.895150298011005\\
-0.499997463456687	0.894599038767121\\
};
\addplot[only marks,mark=x,mark options={},mark size=1.5000pt,draw=black,fill=black] plot table[row sep=crcr]{%
0	1\\
};
\addplot[only marks,mark=*,mark options={},mark size=1.5000pt,draw=black,fill=black] plot table[row sep=crcr]{%
0	0\\
};
\end{axis}
\end{tikzpicture}%

%% file: sensorConeProx.tex
%
%
\definecolor{mycolor1}{rgb}{0.00000,0.44700,0.74100}%
\begin{tikzpicture}

\begin{axis}[%
width=0.75\figurewidth,
height=\figureheight,
at={(0\figurewidth,0\figureheight)},
scale only axis,
unbounded coords=jump,
xmin=-0.7,
xmax=0.7,
xtick={-0.6,-0.4,-0.2,0,0.2,0.4,0.6},
xticklabels={\empty},
xmajorgrids,
ymin=0,
ymax=1.4,
ytick={0,0.2,0.4,0.6,0.8,1,1.2,1.4},
yticklabels={\empty},
ymajorgrids,
axis background/.style={fill=white}
]
\addplot [color=mycolor1,solid,forget plot]
  table[row sep=crcr]{%
1	nan\\
};
\addplot [color=black,solid,forget plot]
  table[row sep=crcr]{%
-0.0499999921748904	0\\
0.0756338867002106	0.0011941236402043\\
0.127611770816773	0.00403052739628644\\
0.167396529532811	0.00793394861912713\\
0.200807304705016	0.0126973794752063\\
0.230088288169689	0.018199715819624\\
0.256383831540301	0.0243579231948942\\
0.280368896408016	0.0311098275105215\\
0.302479540323227	0.0384061246991841\\
0.32301569774975	0.0462060870526444\\
0.34219335019528	0.0544750213469503\\
0.360173564103001	0.0631826557829508\\
0.377079814197621	0.0723020612586533\\
0.393008894425071	0.0818088997940183\\
0.408038096985036	0.0916808833793292\\
0.422230108934574	0.101897373697103\\
0.43563645431468	0.112439079341207\\
0.448299976775065	0.123287822430972\\
0.460256670229628	0.134426355817055\\
0.471537055030832	0.145838217944473\\
0.482167230169537	0.157507616258676\\
0.492169689934896	0.169419332595536\\
0.501563966303543	0.181558645746049\\
0.510367140352844	0.193911267610352\\
0.518594253837726	0.206463290227882\\
0.526258643687932	0.21920114160269\\
0.533372216298598	0.232111548708104\\
0.539945674290044	0.245181506402111\\
0.545988705374943	0.258398251247097\\
0.551510140742605	0.271749239428142\\
0.556518088715224	0.28522212811896\\
0.561020048187836	0.298804759765455\\
0.565023005420156	0.312485148852025\\
0.568533517025208	0.326251470791313\\
0.571557781440182	0.340092052638559\\
0.574101700728559	0.353995365380466\\
0.576170934219397	0.367950017588029\\
0.577770945217853	0.381944750255128\\
0.578907041804222	0.395968432671223\\
0.579584412564662	0.410010059198451\\
0.579808157956171	0.424058746841669\\
0.579583317894096	0.438103733515222\\
0.578914896057062	0.452134376923017\\
0.57780788132756	0.466140153979228\\
0.576267266723085	0.480110660706151\\
0.574298066120265	0.494035612553416\\
0.57190532903067	0.50790484508946\\
0.569094153650421	0.521708315021802\\
0.565869698375008	0.535436101507557\\
0.562237191944825	0.549078407719862\\
0.558201942364995	0.5626255626395\\
0.553769344724491	0.576068023044233\\
0.548944888023649	0.589396375671073\\
0.543734161105688	0.602601339529176\\
0.53814285777616	0.615673768343145\\
0.532176781184334	0.628604653108406\\
0.525841847531864	0.641385124741958\\
0.519144089166667	0.654006456813276\\
0.512089657113483	0.66646006834141\\
0.504684823086988	0.678737526645499\\
0.496935981028448	0.690830550236937\\
0.48884964820265	0.702731011742345\\
0.480432465888107	0.714430940847326\\
0.47169119969031	0.725922527251741\\
0.462632739504876	0.737198123627882\\
0.453264099154981	0.748250248573547\\
0.443592415725206	0.759071589552577\\
0.433624948611985	0.769655005815891\\
0.423369078309102	0.779993531296532\\
0.412832304945159	0.790080377472653\\
0.402022246588561	0.79990893619275\\
0.390946637334382	0.809472782457806\\
0.379613325186383	0.818765677155337\\
0.368030269746503	0.827781569740635\\
0.356205539723287	0.836514600860786\\
0.344147310269982	0.844959104917292\\
0.331863860162315	0.853109612563394\\
0.319363568825397	0.860960853132403\\
0.306654913218629	0.868507756993561\\
0.293746464587028	0.87574545783219\\
0.280646885086921	0.882669294851038\\
0.267364924293616	0.889274814889943\\
0.253909415598255	0.895557774461109\\
0.2402892725008	0.901514141697443\\
0.226513484805766	0.907140098211573\\
0.212591114727111	0.912432040863322\\
0.19853129290845	0.917386583433554\\
0.184343214364554	0.92200055820246\\
0.170036134349924	0.926271017430484\\
0.155619364160074	0.93019523474023\\
0.141102266871003	0.93377070639782\\
0.126494253022223	0.9369951524923\\
0.111804776248579	0.939866518011828\\
0.0970433288660276	0.942382973815471\\
0.0822194374164184	0.944542917499604\\
0.067342658176286	0.946344974157984\\
0.0524225726345597	0.947787997034709\\
0.037468782944072	0.948871068069378\\
0.0224909073516845	0.949593498333895\\
0.00749857561182256	0.949954828360472\\
-0.00749857561182265	0.949954828360472\\
-0.0224909073516844	0.949593498333895\\
-0.0374687829440719	0.948871068069378\\
-0.0524225726345596	0.947787997034709\\
-0.0673426581762861	0.946344974157984\\
-0.0822194374164185	0.944542917499604\\
-0.0970433288660275	0.942382973815471\\
-0.111804776248579	0.939866518011828\\
-0.126494253022223	0.9369951524923\\
-0.141102266871003	0.93377070639782\\
-0.155619364160074	0.93019523474023\\
-0.170036134349924	0.926271017430484\\
-0.184343214364554	0.92200055820246\\
-0.19853129290845	0.917386583433554\\
-0.212591114727111	0.912432040863322\\
-0.226513484805766	0.907140098211573\\
-0.240289272500801	0.901514141697443\\
-0.253909415598255	0.895557774461109\\
-0.267364924293616	0.889274814889943\\
-0.280646885086921	0.882669294851038\\
-0.293746464587028	0.87574545783219\\
-0.306654913218629	0.868507756993561\\
-0.319363568825396	0.860960853132403\\
-0.331863860162315	0.853109612563394\\
-0.344147310269982	0.844959104917292\\
-0.356205539723287	0.836514600860786\\
-0.368030269746503	0.827781569740635\\
-0.379613325186383	0.818765677155337\\
-0.390946637334381	0.809472782457806\\
-0.40202224658856	0.79990893619275\\
-0.412832304945159	0.790080377472653\\
-0.423369078309103	0.779993531296532\\
-0.433624948611984	0.769655005815891\\
-0.443592415725206	0.759071589552577\\
-0.453264099154981	0.748250248573547\\
-0.462632739504876	0.737198123627882\\
-0.47169119969031	0.725922527251741\\
-0.480432465888107	0.714430940847326\\
-0.48884964820265	0.702731011742345\\
-0.496935981028448	0.690830550236937\\
-0.504684823086988	0.678737526645499\\
-0.512089657113483	0.66646006834141\\
-0.519144089166667	0.654006456813276\\
-0.525841847531864	0.641385124741958\\
-0.532176781184334	0.628604653108406\\
-0.53814285777616	0.615673768343145\\
-0.543734161105688	0.602601339529176\\
-0.548944888023649	0.589396375671073\\
-0.553769344724491	0.576068023044233\\
-0.558201942364995	0.5626255626395\\
-0.562237191944825	0.549078407719862\\
-0.565869698375008	0.535436101507557\\
-0.569094153650421	0.521708315021802\\
-0.57190532903067	0.50790484508946\\
-0.574298066120266	0.494035612553416\\
-0.576267266723085	0.480110660706151\\
-0.57780788132756	0.466140153979229\\
-0.578914896057062	0.452134376923017\\
-0.579583317894096	0.438103733515222\\
-0.579808157956171	0.424058746841669\\
-0.579584412564662	0.410010059198451\\
-0.578907041804222	0.395968432671224\\
-0.577770945217853	0.381944750255128\\
-0.576170934219397	0.367950017588029\\
-0.574101700728559	0.353995365380466\\
-0.571557781440182	0.34009205263856\\
-0.568533517025208	0.326251470791313\\
-0.565023005420156	0.312485148852025\\
-0.561020048187836	0.298804759765455\\
-0.556518088715224	0.28522212811896\\
-0.551510140742605	0.271749239428142\\
-0.545988705374943	0.258398251247097\\
-0.539945674290044	0.245181506402111\\
-0.533372216298598	0.232111548708105\\
-0.526258643687932	0.21920114160269\\
-0.518594253837726	0.206463290227882\\
-0.510367140352844	0.193911267610352\\
-0.501563966303543	0.181558645746049\\
-0.492169689934896	0.169419332595536\\
-0.482167230169537	0.157507616258676\\
-0.471537055030832	0.145838217944474\\
-0.460256670229628	0.134426355817055\\
-0.448299976775065	0.123287822430973\\
-0.43563645431468	0.112439079341208\\
-0.422230108934574	0.101897373697103\\
-0.408038096985036	0.0916808833793294\\
-0.393008894425071	0.0818088997940184\\
-0.377079814197621	0.0723020612586533\\
-0.360173564103001	0.0631826557829508\\
-0.34219335019528	0.0544750213469503\\
-0.32301569774975	0.0462060870526443\\
-0.302479540323227	0.0384061246991841\\
-0.280368896408016	0.0311098275105216\\
-0.256383831540301	0.0243579231948943\\
-0.230088288169689	0.018199715819624\\
-0.200807304705017	0.0126973794752064\\
-0.167396529532811	0.00793394861912715\\
-0.127611770816774	0.00403052739628655\\
-0.0756338867002097	0.00119412364020428\\
nan	nan\\
};
\addplot [color=black,solid,forget plot]
  table[row sep=crcr]{%
0.0500000078251096	0\\
0.175621425651976	0.00277274784156799\\
0.22756192972939	0.00718738237276748\\
0.267284398731255	0.0126682476162829\\
0.300607990038166	0.0190079426064736\\
0.329776917214469	0.0260849703603716\\
0.355935559800411	0.0338159039743787\\
0.379758913505705	0.042138177401649\\
0.401683076182552	0.0510020952059558\\
0.422008028769806	0.0603665389951395\\
0.440949805411833	0.0701964256436754\\
0.458669531337012	0.0804610943303608\\
0.475290746189119	0.0911332279076251\\
0.490910314950843	0.102188101423034\\
0.505605606958613	0.113603041064098\\
0.519439392487	0.12535702397634\\
0.532463284854531	0.137430375578238\\
0.544720223026226	0.14980453627084\\
0.556246302245314	0.162461878738756\\
0.567072150182351	0.17538556290794\\
0.577223979108315	0.188559419446093\\
0.586724402526263	0.201967855246557\\
0.59559307753081	0.215595776086461\\
0.603847216190204	0.229428522873695\\
0.611501997090499	0.243451818769537\\
0.618570899798718	0.257651725105891\\
0.625065979117877	0.272014604481339\\
0.63099809180979	0.286527089767325\\
0.636377085423825	0.301176058018118\\
0.641211956641351	0.3159486084787\\
0.645510984890742	0.330832044039662\\
0.649281845744691	0.345813855609056\\
0.652531707668015	0.360881708966309\\
0.655267314960879	0.376023433738881\\
0.657495059182873	0.391227014202792\\
0.659221040907009	0.406480581656913\\
0.660451123309532	0.421772408160457\\
0.661190978829645	0.437090901455439\\
0.66144612991642	0.452424600922415\\
0.661221984706102	0.46776217443981\\
0.66052386833237	0.483092416035322\\
0.659357050457848	0.498404244233184\\
0.657726769521781	0.513686701013835\\
0.655638254122113	0.528928951313314\\
0.653096741886882	0.544120282998862\\
0.65010749613739	0.559250107264935\\
0.646675820601841	0.574307959400503\\
0.642807072401599	0.589283499884163\\
0.638506673501472	0.604166515768476\\
0.633780120789562	0.618946922319193\\
0.628632994930263	0.633614764878634\\
0.623070968115424	0.648160220925705\\
0.617099810822812	0.662573602307781\\
0.610725397677473	0.676845357622105\\
0.603953712499964	0.690966074726483\\
0.596790852615455	0.704926483360923\\
0.589243032489088	0.718717457863492\\
0.581316586745517	0.732330019965165\\
0.573017972624134	0.745755341649688\\
0.564353771915873	0.758984748065657\\
0.555330692422579	0.772009720479039\\
0.545955568975719	0.784821899255271\\
0.536235364047451	0.797413086860899\\
0.526177167983817	0.809775250875479\\
0.515788198886966	0.821900527005112\\
0.505075802170789	0.833781222089588\\
0.494047449812124	0.8454098170957\\
0.482710739317744	0.856778970089737\\
0.471073392425591	0.867881519182678\\
0.459143253557194	0.878710485441974\\
0.446928288036853	0.889259075764238\\
0.434436580091961	0.899520685703482\\
0.421676330647762	0.909488902249885\\
0.408655854928882	0.919157506554379\\
0.395383579879136	0.928520476594603\\
0.381868041410339	0.937571989778076\\
0.368117881490177	0.946306425478638\\
0.354141845078595	0.954718367502485\\
0.339948776921606	0.962802606480304\\
0.325547618210952	0.970554142182241\\
0.310947403117609	0.977968185752621\\
0.296157255206732	0.985040161861536\\
0.281186383741316	0.991765710770555\\
0.26604407988149	0.998140690310045\\
0.250739712786144	1.00416117776567\\
0.23528272562327	1.00982347167188\\
0.219682631495232	1.01512409351022\\
0.203949009284952	1.0200597893107\\
0.188091499428814	1.02462753115412\\
0.172119799621953	1.02882451857403\\
0.156043660461429	1.03264817985652\\
0.139872881032671	1.03609617323658\\
0.12361730444447	1.0391663879897\\
0.107286813317689	1.04185694541754\\
0.0908913252327734	1.0441661997267\\
0.0744407881410876	1.04609273879962\\
0.0579451757450105	1.04763538485675\\
0.0414144828516973	1.04879319500951\\
0.0248587207053469	1.04956546170324\\
0.00828791230279388	1.04995171304989\\
-0.00828791230279398	1.04995171304989\\
-0.0248587207053468	1.04956546170324\\
-0.0414144828516971	1.04879319500951\\
-0.0579451757450104	1.04763538485675\\
-0.0744407881410877	1.04609273879962\\
-0.0908913252327735	1.0441661997267\\
-0.107286813317689	1.04185694541754\\
-0.12361730444447	1.0391663879897\\
-0.139872881032671	1.03609617323658\\
-0.156043660461429	1.03264817985652\\
-0.172119799621954	1.02882451857403\\
-0.188091499428814	1.02462753115412\\
-0.203949009284952	1.0200597893107\\
-0.219682631495232	1.01512409351022\\
-0.23528272562327	1.00982347167188\\
-0.250739712786144	1.00416117776567\\
-0.26604407988149	0.998140690310045\\
-0.281186383741315	0.991765710770555\\
-0.296157255206732	0.985040161861536\\
-0.310947403117609	0.977968185752621\\
-0.325547618210952	0.970554142182241\\
-0.339948776921605	0.962802606480304\\
-0.354141845078595	0.954718367502485\\
-0.368117881490177	0.946306425478638\\
-0.381868041410339	0.937571989778076\\
-0.395383579879136	0.928520476594603\\
-0.408655854928882	0.919157506554379\\
-0.421676330647762	0.909488902249885\\
-0.434436580091961	0.899520685703482\\
-0.446928288036852	0.889259075764238\\
-0.459143253557194	0.878710485441974\\
-0.471073392425592	0.867881519182678\\
-0.482710739317744	0.856778970089737\\
-0.494047449812123	0.8454098170957\\
-0.505075802170788	0.833781222089588\\
-0.515788198886966	0.821900527005112\\
-0.526177167983818	0.809775250875479\\
-0.536235364047451	0.797413086860899\\
-0.545955568975719	0.784821899255271\\
-0.555330692422579	0.77200972047904\\
-0.564353771915873	0.758984748065657\\
-0.573017972624134	0.745755341649688\\
-0.581316586745516	0.732330019965165\\
-0.589243032489088	0.718717457863492\\
-0.596790852615455	0.704926483360923\\
-0.603953712499964	0.690966074726483\\
-0.610725397677473	0.676845357622105\\
-0.617099810822812	0.662573602307781\\
-0.623070968115424	0.648160220925705\\
-0.628632994930263	0.633614764878634\\
-0.633780120789562	0.618946922319193\\
-0.638506673501472	0.604166515768476\\
-0.642807072401599	0.589283499884163\\
-0.646675820601841	0.574307959400503\\
-0.65010749613739	0.559250107264935\\
-0.653096741886882	0.544120282998862\\
-0.655638254122113	0.528928951313314\\
-0.657726769521781	0.513686701013835\\
-0.659357050457848	0.498404244233184\\
-0.66052386833237	0.483092416035322\\
-0.661221984706102	0.46776217443981\\
-0.66144612991642	0.452424600922415\\
-0.661190978829645	0.437090901455439\\
-0.660451123309532	0.421772408160458\\
-0.659221040907009	0.406480581656913\\
-0.657495059182874	0.391227014202792\\
-0.655267314960879	0.376023433738881\\
-0.652531707668015	0.360881708966309\\
-0.64928184574469	0.345813855609056\\
-0.645510984890742	0.330832044039662\\
-0.641211956641351	0.3159486084787\\
-0.636377085423825	0.301176058018118\\
-0.63099809180979	0.286527089767325\\
-0.625065979117877	0.272014604481339\\
-0.618570899798718	0.257651725105891\\
-0.611501997090499	0.243451818769537\\
-0.603847216190204	0.229428522873695\\
-0.59559307753081	0.215595776086461\\
-0.586724402526263	0.201967855246557\\
-0.577223979108315	0.188559419446093\\
-0.567072150182351	0.175385562907941\\
-0.556246302245314	0.162461878738756\\
-0.544720223026226	0.14980453627084\\
-0.532463284854531	0.137430375578238\\
-0.519439392487	0.12535702397634\\
-0.505605606958614	0.113603041064099\\
-0.490910314950843	0.102188101423034\\
-0.475290746189119	0.0911332279076251\\
-0.458669531337012	0.0804610943303607\\
-0.440949805411833	0.0701964256436754\\
-0.422008028769806	0.0603665389951395\\
-0.401683076182552	0.0510020952059559\\
-0.379758913505705	0.0421381774016491\\
-0.355935559800411	0.0338159039743788\\
-0.329776917214469	0.0260849703603717\\
-0.300607990038166	0.0190079426064738\\
-0.267284398731255	0.012668247616283\\
-0.227561929729391	0.00718738237276765\\
-0.175621425651975	0.00277274784156796\\
-0.0500000078251096	6.12323495403631e-18\\
};
\addplot[only marks,mark=x,mark options={},mark size=1.5000pt,draw=black,fill=black] plot table[row sep=crcr]{%
0	1\\
};
\addplot[only marks,mark=*,mark options={},mark size=1.5000pt,draw=black,fill=black] plot table[row sep=crcr]{%
0	0\\
};
\end{axis}
\end{tikzpicture}%

%% file: root.bbl
\begin{thebibliography}{10}
\providecommand{\url}[1]{#1}
\csname url@samestyle\endcsname
\providecommand{\newblock}{\relax}
\providecommand{\bibinfo}[2]{#2}
\providecommand{\BIBentrySTDinterwordspacing}{\spaceskip=0pt\relax}
\providecommand{\BIBentryALTinterwordstretchfactor}{4}
\providecommand{\BIBentryALTinterwordspacing}{\spaceskip=\fontdimen2\font plus
\BIBentryALTinterwordstretchfactor\fontdimen3\font minus
  \fontdimen4\font\relax}
\providecommand{\BIBforeignlanguage}[2]{{%
\expandafter\ifx\csname l@#1\endcsname\relax
\typeout{** WARNING: IEEEtran.bst: No hyphenation pattern has been}%
\typeout{** loaded for the language `#1'. Using the pattern for}%
\typeout{** the default language instead.}%
\else
\language=\csname l@#1\endcsname
\fi
#2}}
\providecommand{\BIBdecl}{\relax}
\BIBdecl

\bibitem{Korthals2016}
T.~Korthals, A.~Skiba, and T.~Krause, ``{Evidenzkarten-basierte Sensorfusion
  zur Umfelderkennung und Interpretation in der Ernte},'' in \emph{Informatik
  in der Land-, Forst und Ern{\"{a}}hrungswirtschaft}, 2016, pp. 15--18.

\bibitem{Korthals2016a}
------, ``{Einsatz Event-Basierter Systemarchitektur f{\"{u}}r Erntemaschinen
  zur Elektronischen Umfelderkennung},'' in \emph{74. Tagung
  LAND.TECHNIK}.\hskip 1em plus 0.5em minus 0.4em\relax VDI e.V., 2016.

\bibitem{Kragh2016}
M.~Kragh, P.~Christiansen, T.~Korthals, T.~Jungeblut, H.~Karstoft, and R.~N.
  J{\o}rgensen, ``{Multi-Modal Obstacle Detection and Evaluation of Occupancy
  Grid Mapping in Agriculture},'' in \emph{International Conference on
  Agricultural Engineering}, Aarhus, 2016.

\bibitem{Korthals2017}
T.~Korthals, J.~Exner, T.~Sch{\"{o}}pping, and M.~Hesse, ``{Semantical
  Occupancy Grid Mapping Framework},'' in \emph{European Conference on Mobile
  Robotics}.\hskip 1em plus 0.5em minus 0.4em\relax IEEE, 2017.

\bibitem{Haehnel2004}
D.~H{\"{a}}hnel, ``{Mapping with Mobile Robots},'' Ph.D. dissertation,
  University of Freiburg, 2004.

\bibitem{Thrun2005}
S.~Thrun, W.~Burgard, and D.~Fox, \emph{{Probabilistic Robotics}}.\hskip 1em
  plus 0.5em minus 0.4em\relax Cambridge, Mass.: MIT Press, 2005.

\bibitem{stachniss09}
C.~Stachniss, \emph{{Robotic Mapping and Exploration}}, 2009.

\bibitem{Garcia2008}
R.~Garcia, O.~Aycard, and T.-d. Vu, ``{High Level Sensor Data Fusion for
  Automotive Applications using Occupancy Grids},'' no. Dec., 2008.

\bibitem{Bouzouraa2010}
M.~E. Bouzouraa and U.~Hofmann, ``{Fusion of occupancy grid mapping and model
  based object tracking for driver assistance systems using laser and radar
  sensors},'' in \emph{2010 IEEE Intelligent Vehicles Symposium}, 2010, pp.
  294--300.

\bibitem{Winner2015}
H.~Winner, \emph{{Handbuch Fahrerassistenzsysteme - Grundlagen, Komponenten und
  Systeme f{\"{u}}r aktive Sicherheit und Komfort}}.\hskip 1em plus 0.5em minus
  0.4em\relax Wiesbaden: Vieweg+Teubner Verlag, 2015.

\bibitem{Elfes1990}
A.~Elfes, ``{Occupancy Grids: A Stochastical Spatial Representation for Active
  Robot Perception},'' in \emph{Proceedings of the Sixth Conference on
  Uncertainty in Artificial Intelligence}, 1990.

\bibitem{Konrad2012}
M.~Konrad, D.~Nuss, and K.~Dietmayer, ``{Localization in digital maps for road
  course estimation using grid maps},'' \emph{IEEE Intelligent Vehicles
  Symposium, Proceedings}, pp. 87--92, 2012.

\bibitem{Bertozzi1996}
M.~Bertozzi and a.~Broggi, ``{Real-time lane and obstacle detection on the GOLD
  system},'' \emph{Proceedings of Conference on Intelligent Vehicles}, 1996.

\bibitem{Simond2003}
N.~Simond and P.~Rives, ``{Homography from a vanishing point in urban
  scenes},'' \emph{International Conference on Intelligent Robots and Systems},
  2003.

\bibitem{Shelhamer2017}
E.~Shelhamer, J.~Long, and T.~Darrell, ``{Fully Convolutional Networks for
  Semantic Segmentation},'' \emph{IEEE Transactions on Pattern Analysis and
  Machine Intelligence}, vol.~39, no.~4, pp. 640--651, 2017.

\bibitem{Kohlbrecher2011}
S.~Kohlbrecher, ``{Grid-based occupancy mapping and automatic gaze control for
  soccer playing humanoid robots},'' \emph{{\ldots} Humanoid Soccer Robots
  {\ldots}}, no. October, 2011.

\bibitem{Redmon2016}
J.~Redmon, S.~Divvala, R.~Girshick, and A.~Farhadi, ``{You Only Look Once:
  Unified, Real-Time Object Detection},'' \emph{Cvpr 2016}, 2016.

\bibitem{Shafer1976}
G.~Shafer, \emph{{A Mathematical Theory of Evidence}}.\hskip 1em plus 0.5em
  minus 0.4em\relax Princeton: Princeton University Press, 1976.

\bibitem{Korthals20161}
T.~Korthals, M.~Barther, and S.~Herbrechtsmeier, ``{Occupancy Grid Mapping with
  Highly Uncertain Range Sensors based on Inverse Particle Filters},'' 2016.

\end{thebibliography}
